\documentclass[10pt,journal,compsoc]{IEEEtran}
\usepackage[cmintegrals]{newtxmath}
\usepackage{graphicx}
\usepackage{epsfig}
\usepackage{amsmath}
\usepackage{colortbl}
\usepackage{comment}
\usepackage{color}
\usepackage{multirow}
\usepackage{epstopdf}
\usepackage{algorithm}
\usepackage{algorithmic}
\usepackage{booktabs}
\usepackage{cite}
\def \etal{{\emph{et al}}}
\def \eg{{\emph{e.g.}}}
\def \ie{{\emph{i.e.}}}
\def \etc{{\emph{etc}}}
\usepackage{pifont}
\usepackage{wasysym}
\usepackage{enumitem}
\newlist{todolist}{itemize}{2}
\setlist[todolist]{label=$\square$}

\usepackage{amsfonts}
\usepackage{array}
\usepackage[caption=false,font=normalsize,labelfont=sf,textfont=sf]{subfig}
\usepackage{textcomp}
\usepackage{stfloats}
\usepackage{url}
\usepackage[colorlinks=true,citecolor=blue]{hyperref}
\usepackage{verbatim}
\usepackage{makecell}
\usepackage{tabularx}

\newcolumntype{I}{!{\vrule width 3pt}}
\newlength\savedwidth

\newlength\savewidth
\newcommand\shline{\noalign{\global\savewidth\arrayrulewidth
\global\arrayrulewidth 1.25pt}%
\hline
\noalign{\global\arrayrulewidth\savewidth}}
\usepackage{capt-of}
\usepackage{caption, subcaption, overpic, textpos}
\usepackage[british, english, american]{babel}

\usepackage{newfloat}
\usepackage{listings}
\DeclareCaptionStyle{ruled}{labelfont=normalfont,labelsep=colon,strut=off} % DO NOT CHANGE THIS
\floatstyle{ruled}
\newfloat{listing}{tb}{lst}{}
\floatname{listing}{Listing}

\renewcommand{\paragraph}[1]{\vspace{1.25mm}\noindent\textbf{#1}}

\usepackage{ragged2e}
\newcommand{\app}{\raise.17ex\hbox{$\scriptstyle\sim$}}

\definecolor{deemph}{gray}{0.6}

\definecolor{baselinecolor}{gray}{.9}

\usepackage{colortbl} %彩色表格需要加载的宏包
\usepackage{xcolor}
\usepackage{bbm}
\definecolor{color4}{rgb}{0.94,0.94,1}
\ifCLASSINFOpdf
\else
\fi
\begin{document}
%\captionsetup[figure]{name={Fig.},labelsep=period}
%
\title{Deep Learning-Driven Ultra-High-Definition Image Restoration: A Survey}
%
% author names and IEEE memberships
% note positions of commas and nonbreaking spaces ( ~ ) LaTeX will not break
% a structure at a ~ so this keeps an author's name from being broken across
% two lines.
% use \thanks{} to gain access to the first footnote area
% a separate \thanks must be used for each paragraph as LaTeX2e's \thanks
% was not built to handle multiple paragraphs
%
%\IEEEcompsocitemizethanks is a special \thanks that produces the bulleted
% lists the Computer Society journals use for "first footnote" author
% affiliations. Use \IEEEcompsocthanksitem which works much like \item
% for each affiliation group. When not in compsoc mode,
% \IEEEcompsocitemizethanks becomes like \thanks and
% \IEEEcompsocthanksitem becomes a line break with idention. This
% facilitates dual compilation, although admittedly the differences in the
% desired content of \author between the different types of papers makes a
% one-size-fits-all approach a daunting prospect. For instance, compsoc 
% journal papers have the author affiliations above the "Manuscript
% received ..."  text while in non-compsoc journals this is reversed. Sigh.
%
\author{Liyan Wang,
        Weixiang Zhou,
        Cong Wang,
        Kin-Man Lam,
        Zhixun Su,
        Jinshan Pan
        % <-this % stops a space
\IEEEcompsocitemizethanks{
% \IEEEcompsocthanksitem This work was supported by the Fundamental Research Funds for the Central Universities (No. 30920041109),
% and the General Research Fund No. 15222220 funded by the UGC of Hong Kong.
% \\
\IEEEcompsocthanksitem Liyan Wang, Weixiang Zhou, and Zhixun Su are with the School of Mathematical Sciences, Dalian University of Technology, Dalian, China (E-mail: \{wangliyan, s20201162006\}@mail.dlut.edu.cn, zxsu@dlut.edu.cn).

\IEEEcompsocthanksitem Cong Wang and Kin-man Lam are with Centre for Advances in Reliability and Safety, Hong Kong, China, and also with the Hong Kong Polytechnic University, Hong Kong, China (E-mail: supercong94@gmail.com,
kin.man.lam@polyu.edu.hk).

% \IEEEcompsocthanksitem Kin-Man Lam is with the Department of Electrical and Electronic Engineering, The Hong Kong Polytechnic University (E-mail: kin.man.lam@polyu.edu.hk).

\IEEEcompsocthanksitem Jinshan Pan is with the School of Computer Science and Engineering, Nanjing University of Science and Technology, Nanjing, China (E-mail: sdluran@gmail.com).
\IEEEcompsocthanksitem Corresponding author: Zhixun Su.
% note need leading \protect in front of \\ to get a newline within \thanks as
% \\ is fragile and will error, could use \hfil\break instead.
% E-mail: see http://www.michaelshell.org/contact.html
}% <-this % stops an unwanted space
%\thanks{Manuscript received April 19, 2005; revised August 26, 2015.}
}
\markboth{IEEE Transactions on Pattern Analysis and Machine Intelligence}%
{Shell \MakeLowercase{\textit{et al.}}: Bare Demo of IEEEtran.cls for Computer Society Journals}
% The only time the second header will appear is for the odd numbered pages
% after the title page when using the twoside option.
% 
% *** Note that you probably will NOT want to include the author's ***
% *** name in the headers of peer review papers.                   ***
% You can use \ifCLASSOPTIONpeerreview for conditional compilation here if
% you desire.
% The publisher's ID mark at the bottom of the page is less important with
% Computer Society journal papers as those publications place the marks
% outside of the main text columns and, therefore, unlike regular IEEE
% journals, the available text space is not reduced by their presence.
% If you want to put a publisher's ID mark on the page you can do it like
% this:
%\IEEEpubid{0000--0000/00\$00.00~\copyright~2015 IEEE}
% or like this to get the Computer Society new two part style.
%\IEEEpubid{\makebox[\columnwidth]{\hfill 0000--0000/00/\$00.00~\copyright~2015 IEEE}%
%\hspace{\columnsep}\makebox[\columnwidth]{Published by the IEEE Computer Society\hfill}}
% Remember, if you use this you must call \IEEEpubidadjcol in the second
% column for its text to clear the IEEEpubid mark (Computer Society jorunal
% papers don't need this extra clearance.)
% use for special paper notices
%\IEEEspecialpapernotice{(Invited Paper)}
% for Computer Society papers, we must declare the abstract and index terms
% PRIOR to the title within the \IEEEtitleabstractindextext IEEEtran
% command as these need to go into the title area created by \maketitle.
% As a general rule, do not put math, special symbols or citations
% in the abstract or keywords.
\IEEEtitleabstractindextext{%
\begin{abstract}
\justifying
Ultra-high-definition (UHD) image restoration aims to specifically solve the problem of quality degradation in ultra-high-resolution images.
Recent advancements in this field are predominantly driven by deep learning-based innovations, including enhancements in dataset construction, network architecture, sampling strategies, prior knowledge integration, and loss functions.
In this paper, we systematically review recent progress in UHD image restoration, covering various aspects ranging from dataset construction to algorithm design. This serves as a valuable resource for understanding state-of-the-art developments in the field.
%with the aim of providing a useful literature review for understanding the field.
%
We begin by summarizing degradation models for various image restoration subproblems, such as super-resolution, low-light enhancement, deblurring, dehazing, deraining, and desnowing, and emphasizing the unique challenges of their application to UHD image restoration.
We then highlight existing UHD benchmark datasets and organize the literature according to degradation types and dataset construction methods.
Following this, we showcase major milestones in deep learning-driven UHD image restoration,  reviewing the progression of restoration tasks, technological developments, and evaluations of existing methods.
We further propose a classification framework based on network architectures and sampling strategies, helping to clearly organize existing methods.
Finally, we share insights into the current research landscape and propose directions for further advancements.
A related repository is available at \url{https://github.com/wlydlut/UHD-Image-Restoration-Survey}. 
\end{abstract}

}

% make the title area
\maketitle
% To allow for easy dual compilation without having to reenter the
% abstract/keywords data, the \IEEEtitleabstractindextext text will
% not be used in maketitle, but will appear (i.e., to be "transported")
% here as \IEEEdisplaynontitleabstractindextext when the compsoc 
% or transmag modes are not selected <OR> if conference mode is selected 
% - because all conference papers position the abstract like regular
% papers do.
\IEEEdisplaynontitleabstractindextext
% \IEEEdisplaynontitleabstractindextext has no effect when using
% compsoc or transmag under a non-conference mode.
% For peer review papers, you can put extra information on the cover
% page as needed:
% \ifCLASSOPTIONpeerreview
% \begin{center} \bfseries EDICS Category: 3-BBND \end{center}
% \fi
%
% For peerreview papers, this IEEEtran command inserts a page break and
% creates the second title. It will be ignored for other modes.
\IEEEpeerreviewmaketitle
% \IEEEraisesectionheading{\section{Introduction}\label{sec:introduction}}
% Computer Society journal (but not conference!) papers do something unusual
% with the very first section heading (almost always called "Introduction").
% They place it ABOVE the main text! IEEEtran.cls does not automatically do
% this for you, but you can achieve this effect with the provided
% \IEEEraisesectionheading{} command. Note the need to keep any \label that
% is to refer to the section immediately after \section in the above as
% \IEEEraisesectionheading puts \section within a raised box.
% The very first letter is a 2 line initial drop letter followed
% by the rest of the first word in caps (small caps for compsoc).
% 
% form to use if the first word consists of a single letter:
% \IEEEPARstart{A}{demo} file is ....
% 
% form to use if you need the single drop letter followed by
% normal text (unknown if ever used by the IEEE):
% \IEEEPARstart{A}{}demo file is ....
% 
% Some journals put the first two words in caps:
% \IEEEPARstart{T}{his demo} file is ....
% 
% Here we have the typical use of a "T" for an initial drop letter
% and "HIS" in caps to complete the first word.
% \IEEEPARstart{T}{his} demo file is intended to serve as a ``starter 

\section{Introduction}\label{sec:introduction}
\begin{figure*}[t]
\centering
\begin{center}
\begin{tabular}{c}
\hspace{-2mm}\includegraphics[width=\linewidth]{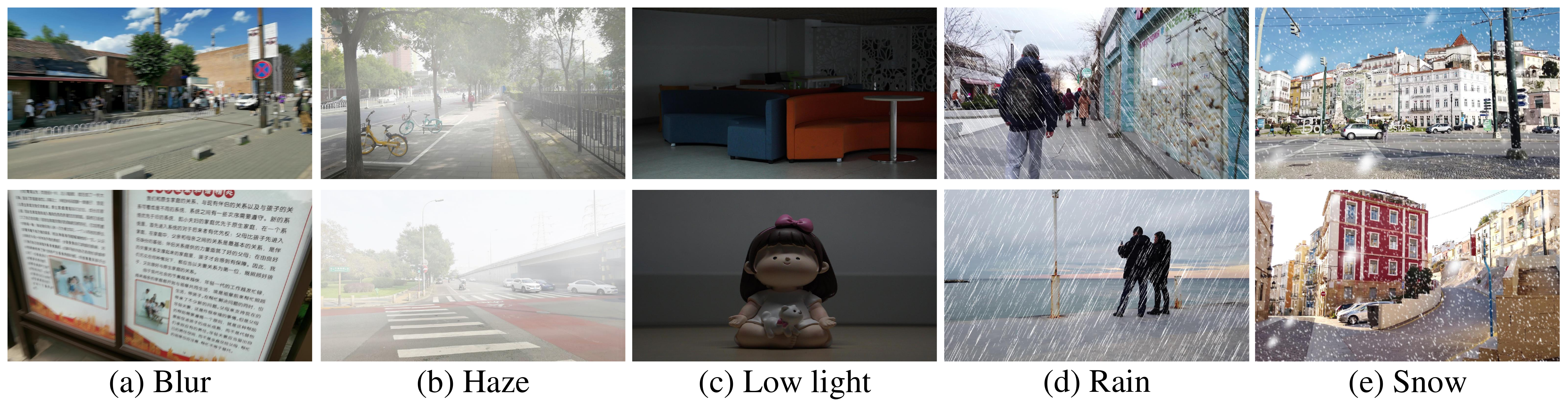} 
\end{tabular}
\caption{Examples of UHD images under different degradation conditions.
These images often suffer from blur, haze, low light, rain, and snow.}\label{fig: different degradation conditions}
\end{center}
\end{figure*}

\IEEEPARstart{R}{ecently}, with the rapid advancement of imaging and acquisition equipment, ultra-high-definition (UHD) images featuring high pixel density and resolutions (\eg, 3,840 $\times$ 2,160 pixels or higher), become widely used in fields such as video streaming~\cite{TANDON2022170164,  Salva-GarciaCAC18}, virtual reality~\cite{ChakareskiA0Z21}, medical imaging~\cite{yamashita2016ultra}, and satellite remote sensing~\cite{Ma23, abs-2411-07688}. This surge in applications has significantly heightened user demand for enhanced image clarity and detail performance.
However, hardware limitations, restricted transmission bandwidth, and challenging acquisition environments often impede the production of high-quality UHD images. Common challenges include insufficient resolution and degradation factors such as blur, rain, snow, haze, low light, \etc. Fig.~\ref{fig: different degradation conditions} illustrates examples of image degradation under different conditions, highlighting how these issues compromise image resolution and overall quality. Consequently, UHD image restoration has emerged as a critical area of research within computer vision and image processing.

UHD image restoration aims to recover high-quality UHD images from degraded inputs, addressing various sub-problems, such as UHD image super-resolution, deblurring, dehazing, low-light image enhancement, deraining, and desnowing. While deep learning-based image restoration methods~\cite{ZamirA0HK0021mprnet,Zamir2021Restormer,0002LZCC21hinet,TuTZYMBL22maxim,wang2022online,wang2024promptrestorer,wang2024msgnn,MeiFZYZLFHS23,ZhangLCGBS22,ChenCZS22nafnet,mirnet20} have achieved remarkable success on lower-resolution images (\eg, low-light image enhancement~\cite{Retinex-Net18, rrdnet22,Wu_2022_CVPR}, dehazing~\cite{wang2024selfpromer, dehazenet16}, desnowing ~\cite{chen2021hdcwnet,ChenFDTK20jstasr}, deraining~\cite{LiWLLZ18, mm20_wang_dcsfn,mm20_wang_jdnet,wang2021single}, and deblurring~\cite{Stripformer,mimo-unet++ }) thanks to advances in techniques like convolutional neural networks (CNNs)~\cite{KrizhevskySH12, HeZRS16} and Transformers~\cite{VaswaniSPUJGKP17, DosovitskiyB0WZ21}, their effectiveness remains limited when applied to UHD image restoration. This limitation stems from the fact that these models are often designed for low-resolution inputs and struggle to scale efficiently to UHD images. In addition, UHD images inherently possess more intricate details, a wider color gamut, and a significantly larger number of pixels, which pose unique computational challenges. 

\begin{figure*}[t]
\centering
\begin{center}
\begin{tabular}{c}
\hspace{-2mm}\includegraphics[width=\linewidth]{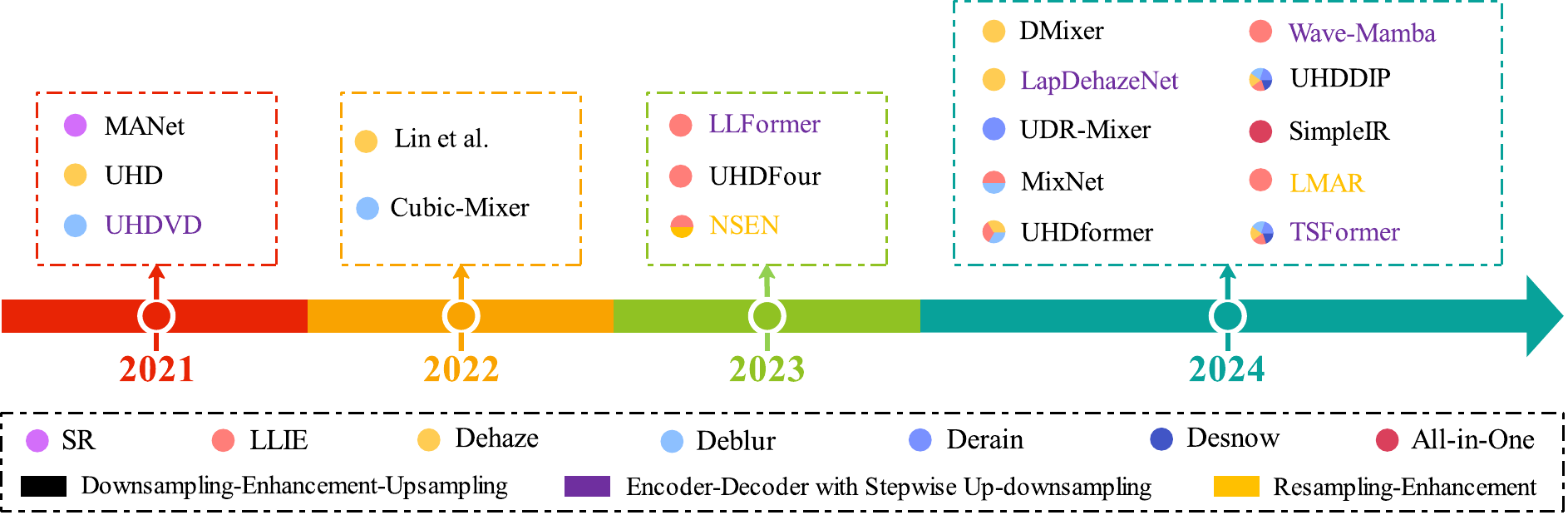} 
\put(-456,126.5){\scriptsize{\cite{ZhangLLRS0L021}}}
\put(-463.5,110.5){\scriptsize{\cite{ZhengRCHWSJ21}}}
\put(-451.5,94.5){\scriptsize{\cite{DengRYWSC21}}}
\put(-356,122){\scriptsize{\cite{LinZJ22}}}
\put(-343,99.2){\scriptsize{\cite{cub-mixer}}}
\put(-246,126.5){\scriptsize{\cite{LLformer}}}
\put(-246,110){\scriptsize{\cite{LiGZLZFL23}}}
\put(-260,94){\scriptsize{\cite{nsen}}}
%%%%%%%%%%%%%%%%%%%%
\put(-152.5,157.8){\scriptsize{\cite{dmixer}}}
\put(-128,141.5){\scriptsize{\cite{LapDehazeNet}}}
\put(-138,125){\scriptsize{\cite{UDRMixer}}}
\put(-152.5,108){\scriptsize{\cite{MixNet}}}
\put(-137,91.5){\scriptsize{\cite{uhdformer}}}
\put(-43,157){\scriptsize{\cite{Wave-Mamba}}}
\put(-58,142.5){\scriptsize{\cite{uhddip}}}
\put(-58,125){\scriptsize{\cite{SimpleIR}}}
\put(-66,108){\scriptsize{\cite{LMAR}}}
\put(-55.5,91.5){\scriptsize{\cite{tsformer}}}
\end{tabular}
\caption{Milestones in UHD image restoration methods are categorized into three primary structures: 
Downsampling-Enhancement-Upsampling, Encoder-Decoder with Stepwise Up-downsampling, and Resampling-Enhancement.}\label{fig: Milestones of UHD image restoration methods}
\end{center}
\end{figure*}
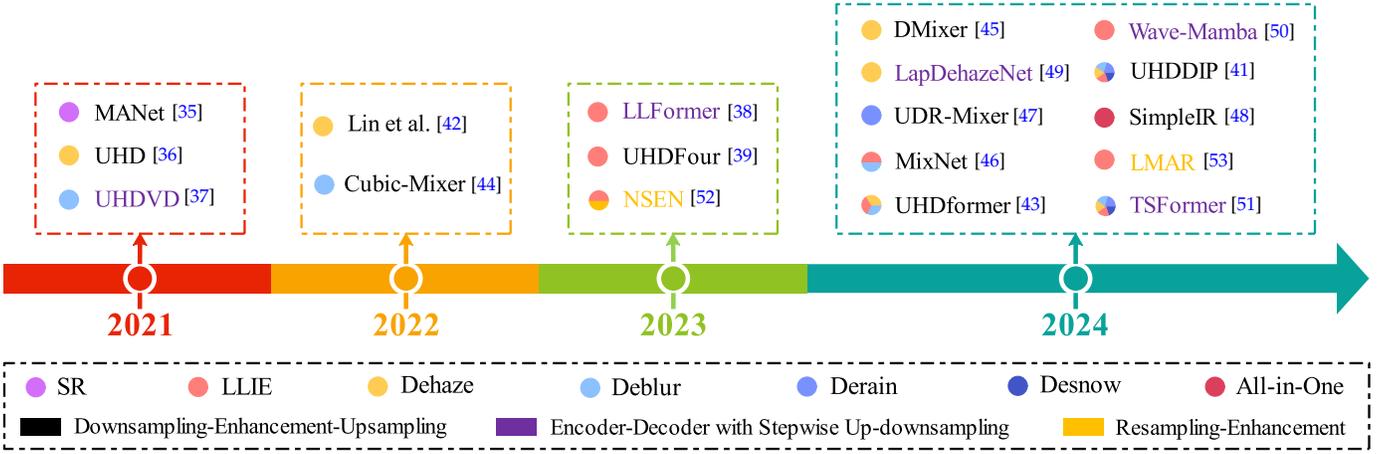

Since the introduction of the first large-scale dataset for UHD image super-resolution reconstruction tasks~\cite{ZhangLLRS0L021} in 2021, a range of UHD image restoration methods and corresponding datasets have been developed. These include multi-guided bilateral learning for UHD image dehazing~\cite{ZhengRCHWSJ21} with the 4KID dataset; multi-scale separable-patch integration networks for video deblurring~\cite{DengRYWSC21} with the 4DRK dataset; the Transformer-based LLFormer~\cite{LLformer} with the UHD-LOL dataset and the Fourier embedding network UHDFour~\cite{LiGZLZFL23} with the UHD-LL dataset for UHD low-light image enhancement; and UHDformer~\cite{wang2024uhdformer}, which explores feature transformation between high- and low-resolution, with the UHD-Haze/UHD-Blur datasets for UHD image restoration. Additionally, the dual interaction prior-driven network UHDDIP~\cite{uhddip}, paired with the UHD-Snow/UHD-Rain datasets, addresses UHD image restoration.
These methods for processing UHD images employ diverse strategies, including downsampling-enhancement-upsampling structure~\cite{DengRYWSC21,LinZJ22,LiGZLZFL23,uhdformer,cub-mixer,dmixer,MixNet,UDRMixer,SimpleIR,uhddip}, encoder-decoder structures with stepwise up-downsampling~\cite{LapDehazeNet,LLformer,Wave-Mamba,tsformer}, and resampling-enhancement structures~\cite{nsen,LMAR}.
The evolution of network models has transitioned from CNNs~\cite{DengRYWSC21,LinZJ22,LiGZLZFL23,LapDehazeNet}, which emphasize local feature extraction, to Transformer-based models~\cite{LLformer,uhdformer,SimpleIR,uhddip}, which focus on global modeling, and most recently to Multilayer Perceptrons (MLPs)~\cite{cub-mixer,dmixer,MixNet,UDRMixer} and Mamba~\cite{Wave-Mamba}, which aim to reduce computational overhead.
From 2021 to the present, approximately 20 studies have explored deep learning-based UHD image restoration methods. A concise summary of these developments is presented in Fig.~\ref{fig: Milestones of UHD image restoration methods}.

Although deep learning has dominated research on UHD image restoration, there is a lack of comprehensive and in-depth surveys on deep learning-based solutions. Therefore, our work systematically and comprehensively reviews the research in the field to provide a useful starting point for understanding the major developments, limitations of existing approaches, and potential future research directions. 
The main contributions of this paper are threefold:
\begin{itemize}
    \item We conduct a systematic review of research progress in deep learning-based UHD image restoration, covering problem definitions for various UHD restoration tasks, challenges, the development of benchmark datasets, and the improvements and limitations of existing methods.
    \item We propose an effective classification method for existing deep learning-based UHD image restoration methods and analyze representative benchmarks under different subtasks.
    \item Finally, we discuss the challenges faced by current deep learning methods and outline promising future directions.
\end{itemize}
The rest of the paper is organized as follows. Section~\ref{Sec: Promlem Formulation} describes the degradation models for the sub-problems of image restoration, including low-resolution, low-light, blur, haze, rain, and snow, as well as the challenges of UHD image restoration. Section~\ref{Sec: UHD-snow and UHD-Rain Datasets} presents a detailed survey of benchmark datasets for UHD image restoration, covering the process of constructing the datasets and benchmark classification. Section~\ref{Sec: Deep Learning-based UHD Image Restoration Approaches} provides a comprehensive survey of UHD image restoration methods, including technological developments in tasks such as single image super-resolution, low-light image enhancement, image deblurring, dehazing, deraining, and desnowing. Section~\ref{Sec: Technical Development Review} reviews the technical development of current methods, including network architectures, sampling strategies, and loss functions. Section~\ref{Sec: Performance Evaluation} presents a quantitative and qualitative comparison of several representative UHD image restoration methods. Finally, we explore potential future research directions in Section~\ref{Sec: Challenges and Opportunities} and conclude with a summary in Section~\ref{Sec: Conclusion}.

\section{Promlem Formulation}\label{Sec: Promlem Formulation}
In this section, we provide a detailed overview of various degradation models associated with image restoration. These models address common scenarios, such as low-resolution, low-light, blur, haze, rain, and snow, and outline the main challenges faced in UHD image restoration.
% \\
\subsection{Image Degradation}
The primary objective of image restoration is to recover a clean image $J(x)$ from a degraded image $I(x)$. The degradation model is mathematically expressed as:
\begin{equation}
\begin{array}{ll}
I(x) = f(J(x))+N,
\end{array}
\label{eq:degradations}
\end{equation}
where $f$ is the function that defines the degradation process, including downsampling, blur, low light, haze, rain, and snow, and $N$ is additive noise. Therefore, image restoration is fundamentally the problem of modeling and solving the inverse function $f$. A detailed description of the various degradation models excluding noise, is provided below.
\\
\noindent\textbf{Super-Resolution.}\; Image super-resolution (SR) degradation assumes that a high-resolution (HR) image undergoes Gaussian blur followed by downsampling resulting in a low-resolution (LR) image and the consequent loss of details. The SR degradation process can be expressed as follows:
\begin{equation}
\begin{array}{ll}
I_{LR}(x) = (J_{HR}(x)\otimes K)\downarrow_S,
\end{array}
\label{eq:sr_degradation}
\end{equation}
where $\otimes$ represents the convolution operation, $K$ is the Gaussian blur kernel, and $\downarrow_S$ denotes downsampling by a scale factor $S$.
The LR image $I_{LR}(x)$ is derived from the HR image $J_{HR}(x)$ using this process.
\\
\noindent\textbf{Blur.}\; Blurred images often arise from camera or object movement and focus failure, manifesting as motion blur and focus blur. This degradation can be modeled as follows:
\begin{equation}
\begin{array}{ll}
I_{B}(x) = J(x)\otimes K,
\end{array}
\label{eq:blur_degradation}
\end{equation}
where $K$ is the Gaussian blur kernel, and $I_{B}(x)$ is represents the blurred image.
\\
\noindent\textbf{Low-Light.}\; According to the Retinex model, an image can be decomposed into a reflection component and an illumination component. Under the assumption that the image content is independent of the illumination component, a low-light image $I_{LL}(x)$ can be generated as follows:
\begin{equation}
\begin{array}{ll}
I_{LL}(x)=J(x)L,
\end{array}
\label{eq:Lowlight_degradation}
\end{equation}
where $L$ is a random illumination value in the range of $[0, 1]$.
\\
\noindent\textbf{Haze.}\; The generation of haze images $I_{H}(x)$ is usually represented using the atmospheric scattering model:
\begin{equation}
\begin{array}{ll}
I_{H}(x)=J(x)t(x) + A(1-t(x)),
\end{array}
\label{eq:haze_degradation}
\end{equation}
where $A$ denotes atmospheric light, and $t(x) $ is the transmission map, defined as $t(x) = e^{-\beta d(x)}$, where $\beta$ represents the haze density coefficient and $d(x)$ is the scene depth.
\\
\noindent\textbf{Rain/Snow.}\; The impact of rain and snow can be modeled using an additive composite model, as follows:
\begin{equation}
\begin{array}{ll}
I_{RS}(x)=J(x)+R,
\end{array}
\label{eq:rainsnow_degradation}
\end{equation}
where $I_{RS}(x)$ and $J(X)$ represent a rain or snow image and the corresponding clean image, respectively,
and $R$ denotes the streaks introduced by rain and snow.

\subsection{Challenges in UHD Image Restoration}
UHD images, affected by the aforementioned degradation processes, pose unique challenges for restoration due to their ultra-high resolution and dense pixel characteristics.
\begin{itemize}
    \item \textbf{High computational requirements.}\; Compared to high-definition (HD) images, UHD images contain significantly more pixels. Processing such large-scale feature maps demands substantial computational resources and storage necessitating advanced GPUs and high-performance hardware.
   
    \item \textbf{Difficulty in recovering details.}\; UHD images capture intricate details, and degradation effects are more pronounced at high resolutions. Although existing methods perform adequately with low-resolution images, they often struggle to effectively preserve fine structures in UHD images, resulting in texture loss or artifacts in the restoration output.
   
    \item \textbf{Lack of training datasets.}\; Methods trained on low-resolution datasets cannot directly process UHD images, as they require large amounts of degraded-clear UHD image pairs for fine-tuning. However, most publicly available datasets are dominated by HD and lower-resolution images, with a notable lack of datasets specialized for UHD scenarios. This limitation significantly hinders the development and testing of algorithms.
    
    \item \textbf{Poor algorithm adaptability.}\; Many restoration algorithms are designed for low-resolution images and fail to scale effectively to UHD images. Developing new algorithms that accommodate the distinct characteristics of UHD images is an urgent need.
\end{itemize}

\section{Benchmark Datasets for UHD Image Restoration }\label{Sec: UHD-snow and UHD-Rain Datasets}
This section highlights the key benchmark datasets developed for UHD image restoration, categorized by the specific challenges they address.
\begin{table*}[t]
\setlength{\tabcolsep}{9.5pt}
\caption{Representative benchmark datasets for UHD image restoration in previous works.}\label{tab:Benchmark Datasets}
\centering
\renewcommand{\arraystretch}{1.4}
\begin{tabular}{l|c|cccc}
\shline
\textbf{Datasets} &\textbf{Restoration Task} &  \textbf{Number (Train / Test)} &  \textbf{Synthetic / Real} &  \textbf{Format}  &\textbf{Publication} 
\\
\shline
\textbf{UHDSR4K}~\cite{ZhangLLRS0L021} &Image Super-Resolution &5,999 / 2,100 &Real &PNG &ICCV’21\\
\shline
\textbf{4KRD}~\cite{DengRYWSC21}&Video Deblurring &19,958 / 1600  &Synthetic &JPG &ICCV’21\\
\textbf{UHD-Blur}~\cite{wang2024uhdformer}&Image Deblurring&1,964 / 300 &Synthetic &JPG & AAAI’24 \\
\shline
\textbf{4KID}~\cite{ZhengRCHWSJ21}&\multirow{3}{*}{Image Dehazing} &8,000 / 200 &Synthetic &JPG & CVPR’21\\
\textbf{Updated 4KID}~\cite{XiaoZZLJ24}& &10,400 / 200 &Synthetic &JPG &Signal Process.’24\\
\textbf{UHD-Haze}~\cite{wang2024uhdformer}&&2,290 / 230 &Synthetic &JPG &AAAI’24 \\
\shline
\textbf{4KIL}~\cite{LinZJ22}&\multirow{3}{*}{Low-Light Image Enhancing} &1,000 / 100 &Synthetic &JPG & Inf. Sci’22\\
\textbf{UHD-LOL4K}~\cite{LLformer}& &5,999 / 2,100 &Synthetic &PNG &  AAAI’23  \\
\textbf{UHD-LL}~\cite{LiGZLZFL23}&&2,000 / 150  &Real &JPG &ICLR'23 \\
\shline
\textbf{4K-Rain13K}~\cite{UDRMixer}&\multirow{3}{*}{Image Deraining}&12,500 / 500&Synthetic &JPG &ArXiv’24 \\
\textbf{4K-RealRain}~\cite{UDRMixer}&&- / 320&Real &JPG &ArXiv’24 \\
\textbf{UHD-Rain}~\cite{uhddip} & &3,000 / 200&Synthetic &JPG &ArXiv’24\\
\shline
\textbf{UHD-Snow}~\cite{uhddip} &Image Desnowing &3,000 / 200 &Synthetic &JPG &ArXiv’24 \\
\shline
\end{tabular}
\end{table*}

\subsection{UHD Image Super-Resolution Datasets}\label{Sec:UHD Image Super-Resolution Datasets}
\noindent\textbf{UHDSR4K Dataset.}\; To support the evaluation of image super-resolution methods, Zhang \etal.~\cite{ZhangLLRS0L021} introduce the UHDSR4K dataset, the first large-scale 4K (3,840 × 2,160 pixels) image super-resolution dataset. The original 4K images are collected from the Internet and captured by various devices, featuring a diverse range of indoor and outdoor scenes, such as cityscapes, people, animals, buildings, cars, natural landscapes, and sculptures. 

The UHDSR4K dataset is divided into two subsets: a training set consisting of 5,999 HR images and a testing set containing 2,100 HR images.
LR images are synthesized from the HR images by simulating degradation~\cite{GuLDFLT19,TimofteAG0ZLSKN17}. Specifically, seven types of degradation are applied, including downsampling by factors of $2\times$, $3\times$, $4\times$, $8\times$, $16\times$, blurring followed by $3\times$ downsampling, and $3\times$ downsampling followed by the addition of Gaussian noise with $\sigma=30$. Gaussian blurring is conducted using a kernel size of $7\times7$ and a standard deviation of $1.6$. 

\subsection{UHD Low-Light Image Enhancing Datasets}\label{Sec:UHD Low-Light Image Enhancing Datasets}
\noindent\textbf{4KIL Dataset.}\; Lin \etal.~\cite{LinZJ22} introduce the 4K low-light image dataset 4KIL, consisting of 10,000 frames of low-light and normal-light images extracted from 100 4K resolution video clips. Among them, 1,000 pairs are designated for training while an additional 100 pairs are for testing. The low-light images are synthesized based on a linear combination and gamma transformation with parameters set as follows: $\beta\sim U(0.5. 1)$, $\alpha\sim U(0.9. 1)$, and $\gamma\sim U(1.5. 5)$.

\noindent\textbf{UHD-LOL4K Dataset.}\; To evaluate low-light image enhancement algorithms, Wang \etal.~\cite{LLformer} synthesize low-light images using high-resolution normal-light images from UHDSR4K~\cite{ZhangLLRS0L021}. 
These images are created using Adobe Lightroom software with the following settings: exposure $(-5+5X^2)$, highlights $(50min\{Y,0.5\}+75)$, shadows $(-100min\{Z,0.5\})$, vibrance $(-75+75X^2)$, and whites $(16(5-5X^2))$.
UHD-LOL4K contains 8,099 pairs of 4K low-light and normal-light images, divided into 5,999 pairs for training and 2,100 pairs for testing.

\noindent\textbf{UHD-LL Dataset.}\; Li \etal.~\cite{LiGZLZFL23} introduce the first paired normal-light and low-light image dataset UHD-LL captured in real-world scenes. This dataset contains 2,150 4K UHD image pairs, with 2,000 pairs for training and 150 pairs for testing. 
The real image pairs are captured using two cameras (a Sony $\alpha$7 III camera and a Sony Alpha a6300 camera) mounted on tripods. The normal-light images are captured using a small $ISO\in[100, 800]$ in bright environments (indoor or outdoor), while the low-light images are generated by increasing the $ISO\in[1000, 20000]$ and reducing the exposure time.

\subsection{UHD Image Deblurring Datasets}\label{Sec:UHD Image Deblurring Datasets}
\noindent\textbf{4KRD Dataset.}\; Deng \etal.~\cite{DengRYWSC21} construct the 4K video deblurring dataset, 4KRD, which includes both synthetic and real captured videos. The captured videos are shot using three different smartphones. A frame interpolation method is then utilized to generate high frame rate 4K videos, increasing the frame rate from 30/60 fps to 480 fps. Similar to~\cite{NahTBHMSL19}, blurry frames are obtained through frame averaging.

\noindent\textbf{UHD-Blur Dataset.}\; Wang \etal.~\cite{wang2024uhdformer} select 2,264 blur/clean image pairs from the 4KRD dataset~\cite{DengRYWSC21} to form a new benchmark, UHD-Blur. This dataset comprises 1,964 pairs for training and 300 pairs for testing. 

\subsection{UHD Image Dehazing Datasets}\label{Sec:UHD Image Dehazing Datasets}
\noindent\textbf{4KID Dataset.}\; To evaluate dehazing methods, Zheng \etal.~\cite{ZhengRCHWSJ21} construct the 4KID UHD image dehazing dataset, containing 10,000 haze/sharp image pairs extracted from 100 video clips at 4K resolution. Among these, 8,000 images are used as training, while 200 images are selected from the remaining data for testing. The haze images are synthesized using the atmospheric scattering model with atmospheric light $A\in[0.8, 1]$ and scattering coefficients $\alpha\in[0.4, 2.0]$.

\noindent\textbf{Updated 4KID Dataset.}\; Xiao \etal.~\cite{XiaoZZLJ24} expand the 4KID dataset introduced by Zheng \etal.~\cite{ZhengRCHWSJ21} by including an additional 3,000 4K images captured using different cell phones. Corresponding haze images are synthesized based on the atmospheric scattering model. In this updated dataset, 10,400 images are used for training and 200 for testing.

\noindent\textbf{UHD-Haze Dataset.}\; Following a similar approach to UHD-Blur, Wang \etal.~\cite{wang2024uhdformer} curate a new benchmark dataset, UHD-Haze, by selecting 2,521 haze/clean image pairs from the 4KID dataset~\cite{ZhengRCHWSJ21}. Out of this, 2,290 pairs are allocated for training and 231 pairs for testing.

\subsection{UHD Image Deraining Datasets}\label{Sec:UHD Image Deraining Datasets}
\noindent\textbf{UHD-Rain Dataset.}\; To evaluate the performance of image deraining algorithms, Wang \etal.~\cite{uhddip} create a synthetic UHD-Rain dataset, containing 3,200 pairs of rain/clear 4K resolution images. Among them, 3,000 pairs are used for training and 200 pairs for testing. The clear images are randomly selected from high-resolution samples in the UHDSR4K dataset~\cite{ZhangLLRS0L021}, and the rain images are then obtained by superimposing rain masks onto the clear images, following the methodology described in~\cite{YangTFGYL20, YangTFLGY17}. 
The rain masks, created using Photoshop, feature 50 different orientations and 4 densities of rain streaks.
The parameter settings include noise $(50\%)$, crystallize $(5)$, motion blur $(200)$, Gaussian blur $(2)$, threshold $(55-67)$, and angle $(45^{\circ}-135^{\circ})$.

\noindent\textbf{4K-Rain13k Dataset.}\; Unlike previous methods~\cite{YangTFGYL20, YangTFLGY17} for synthesizing rain datasets, Chen \etal.~\cite{UDRMixer} synthesize corresponding rain images by modeling rain streak generation as a motion blur process. They further introduce geometric transformations to adjust the scale of the rain streaks in the UHD images. Before this, clear UHD images are downloaded from the web using a Scrapy-based Python program, including daytime and nighttime urban scenes (buildings, streets, cityscapes) as well as natural landscapes (lakes, hills, and vegetation). The 4K-Rain13k dataset ultimately contains 12,500 synthetic training pairs and 500  pairs of testing images.

\noindent\textbf{4K-RealRain Dataset.}\; Chen \etal.~\cite{UDRMixer} also create a collection of 320 real rainy-day images from the Internet and real-world environments captured using smartphones, though without ground-truth images.
\begin{figure*}[t]
\centering
\begin{center}
\begin{tabular}{c}
\hspace{-2mm}\includegraphics[width=\linewidth]{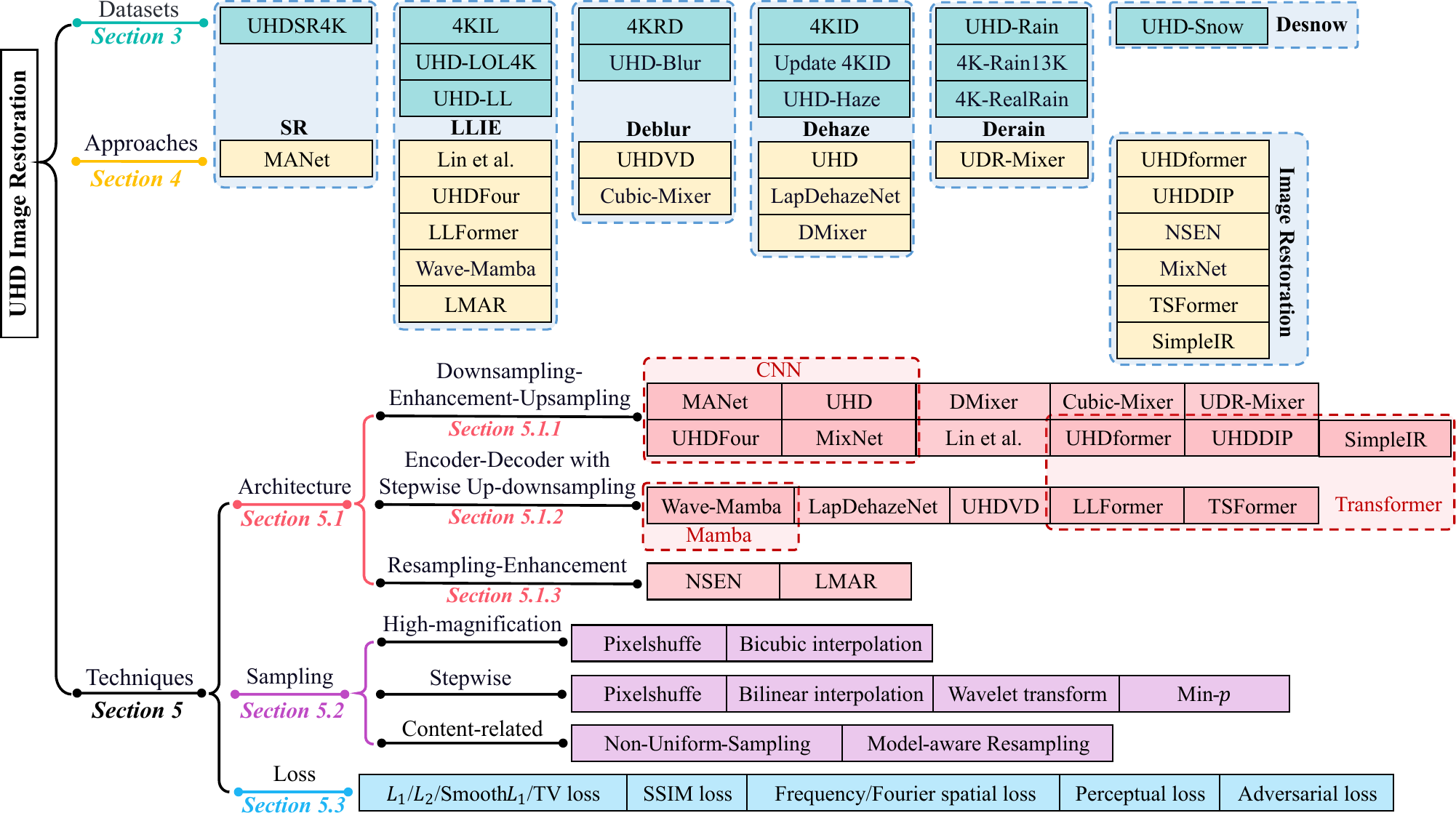} 
\end{tabular}
\caption{Taxonomy of deep learning-based UHD image restoration across datasets, approaches, and techniques.
}\label{fig: xmind}
\end{center}
\end{figure*}
%%%%%%%%%%%%%%%%%%%
\subsection{UHD Image Desnowing Datasets}\label{Sec:UHD Image Desnowing Datasets }
\noindent\textbf{UHD-Snow Dataset.}\; Wang \etal.~\cite{uhddip} develop a new synthetic snow dataset, UHD-Snow, for evaluating UHD image desnowing methods. This dataset leverages UHDSR4K HR images as backgrounds.  
The snow images are created by superimposing snow masks, generated by using Photoshop software, onto the clear images, following CSD~\cite{ChenFHTCDK21CSD}. The snow masks feature snow streaks with 10 different densities and snowflakes of different sizes.
The parameter settings for generating snow masks include noise $(50\%)$, crystallize $(15)$, motion blur $(25)$, Gaussian blur $(5)$, threshold $(100-165)$, and flows $(60\%,100\%)$.
The UHD-Snow dataset contains 3,200 image samples, including 3,000 for the training set and 200 for testing.

\begin{table*}[t]
%\scriptsize
\setlength{\tabcolsep}{3.75pt}
\caption{Summary of essential characteristics of representative deep learning-based UHD image restoration methods.}\label{tab:Deep Learning-Based UHD Image Restoration Methods}
\centering
\renewcommand{\arraystretch}{1.3}
\begin{tabular}{l|c|ccccc}
\shline
\textbf{Methods}&\textbf{Category} &\textbf{Task} & \textbf{Architecture} &  \textbf{Loss Function} & \textbf{Sampling Strategy}& \textbf{Evakuation Metric}
\\
\shline
\cellcolor{gray!20}\textbf{MANet} &\multirow{22}{*}{\makecell[c]{Downsampling-\\Enhancement-\\Upsampling}} &\multirow{2}{*}{SR} &\multirow{2}{*}{CNN} &\multirow{2}{*}{-}&\multirow{2}{*}{-} &\multirow{2}{*}{PSNR/SSIM}  \\
\cellcolor{gray!20}\cite{ZhangLLRS0L021}&&&&&&\\
\textbf{UHD}& &\multirow{2}{*}{Dehaze} &\multirow{2}{*}{CNN} & \multirow{2}{*}{$L_2$ loss}&\multirow{2}{*}{\makecell[c]{High-magnifcation\\Bicubic (256,256) }}&\multirow{2}{*}{PSNR/SSIM} \\
\cite{ZhengRCHWSJ21}&&&&&&\\
\cellcolor{gray!20}
\textbf{Lin \etal.}& &\multirow{2}{*}{LLIE} &\multirow{2}{*}{UNet} &\multirow{2}{*}{\makecell[c]{$L_1$\&Fourier \\ Spatial loss}}&\multirow{2}{*}{\makecell[c]{High-magnifcation\\Bicubic (256,256) }}&\multirow{2}{*}{PSNR/SSIM} \\
\cellcolor{gray!20}\cite{LinZJ22}&&&&&&\\
\textbf{Cubic-Mixer}&&\multirow{2}{*}{Deblur} &\multirow{2}{*}{MLP} &\multirow{2}{*}{$L_1$\&Perceptual
loss} &\multirow{2}{*}{\makecell[c]{High-magnifcation\\Bicubic $\times$16/32/64 }}&\multirow{2}{*}{PSNR/SSIM} \\  \cite{cub-mixer}&&&&&& \\
\cellcolor{gray!20}\textbf{UHDFour}&&\multirow{2}{*}{LLIE}&\multirow{2}{*}{CNN} &$L_1$\&SSIM&\multirow{2}{*}{\makecell[c]{High-magnifcation\\Pixelshuffe $\times$8}} &\multirow{2}{*}{PSNR/SSIM/LPIPS}\\
\cellcolor{gray!20}\cite{LiGZLZFL23}&&&&\&Perceptual loss&& \\
\textbf{DMixer}&&\multirow{2}{*}{Dehaze} &\multirow{2}{*}{MLP} &\multirow{2}{*}{$L_1$ loss} &\multirow{2}{*}{\makecell[c]{High-magnifcation\\Bicubic (384,384)}}&\multirow{2}{*}{PSNR/SSIM/FOM} \\  \cite{dmixer}&&&&&& \\
\cellcolor{gray!20}\textbf{MixNet}&&\multirow{2}{*}{LLIE/Deblur} &\multirow{2}{*}{CNN} &\multirow{2}{*}{$L_1$ loss}&\multirow{2}{*}{\makecell[c]{High-magnifcation\\Pixelshuffe $\times$4 }}  &\multirow{2}{*}{PSNR/SSIM}   \\
\cellcolor{gray!20}\cite{MixNet}&&&&&&  \\
\textbf{UDR-Mixer}&&\multirow{2}{*}{Derain} &\multirow{2}{*}{MLP} &\multirow{2}{*}{$L_1$ loss}&\multirow{2}{*}{\makecell[c]{High-magnifcation\\Pixelshuffe $\times$4}} &PSNR/SSIM/MSE/   \\
\cite{UDRMixer}&&&&&&NIQE/PIQE/BRISQUE\\
\cellcolor{gray!20}
\textbf{UHDformer}&&\multirow{2}{*}{LLIE/Deblur/Dehaze} &\multirow{2}{*}{Transformer} &\multirow{2}{*}{$L_1$\&Frequency
loss} &\multirow{2}{*}{\makecell[c]{High-magnifcation\\Pixelshuffe $\times$8}} &\multirow{2}{*}{PSNR/SSIM} \\  \cellcolor{gray!20}\cite{uhdformer}&&&&&& \\
\textbf{UHDDIP}&&LLIE/Dehaze/Deblur &CNN\& &\multirow{2}{*}{$L_1$\&Frequency
loss} &\multirow{2}{*}{\makecell[c]{High-magnifcation\\Pixelshuffe $\times$8}} &\multirow{2}{*}{PSNR/SSIM/LPIPS}  \\
\cite{uhddip}&&/Derain/Desnow&Transformer&&&\\
%
%\cellcolor{gray!20}\textbf{UHDPromer}&&\multirow{2}{*}{LLIE/Dehaze/Deblur}&\multirow{2}{*}{Transformer }&\multirow{2}{*}{$L_1$\&frequency loss} &\multirow{2}{*}{\makecell[c]{High-magnifcation\\Pixelshuffe $\times$8 }} &\multirow{2}{*}{PSNR/SSIM/LPIPS} \\
%\cellcolor{gray!20}-&&&&&& \\
%
\cellcolor{gray!20}\textbf{SimpleIR}&&\multirow{2}{*}{All-in-One} &\multirow{2}{*}{Transformer} &\multirow{2}{*}{$L_1$ \&Frequency
loss}&\multirow{2}{*}{\makecell[c]{High-magnifcation\\Pixelshuffe $\times$4 }} &\multirow{2}{*}{PSNR/SSIM/LPIPS}   \\
\cellcolor{gray!20}\cite{SimpleIR}&&&&&& \\
\shline
\textbf{UHDVD}&\multirow{10}{*}{\makecell[c]{Encoder-Decoder \\with Stepwise Up\\-downsampling}}&\multirow{2}{*}{Deblur} &\multirow{2}{*}{UNet} & \multirow{2}{*}{$L_2$\&TV loss}&\multirow{2}{*}{\makecell[c]{Stepwise Sampling\\Bilinear}}&\multirow{2}{*}{PSNR/SSIM} \\
\cite{DengRYWSC21}&&&&&&\\
\cellcolor{gray!20}\textbf{LLFormer}&&\multirow{2}{*}{LLIE} &\multirow{2}{*}{Transformer} &\multirow{2}{*}{Smooth $L_1$ loss}&\multirow{2}{*}{\makecell[c]{Stepwise Sampling\\Pixelshuffle}} &PSNR/SSIM/LPIPS/ \\
\cellcolor{gray!20}\cite{LLformer}&&&&&&MAE\\
\textbf{LapDehazeNet}& &\multirow{2}{*}{Dehaze} &\multirow{2}{*}{UNet} & \multirow{2}{*}{Charbonnier loss}&\multirow{2}{*}{\makecell[c]{Stepwise Sampling\\Bilinear}}&\multirow{2}{*}{PSNR/SSIM} \\
\cite{LapDehazeNet}&&&&&&\\
\cellcolor{gray!20}\textbf{Wave-Mamba}&&\multirow{2}{*}{LLIE} &\multirow{2}{*}{Mamba} &\multirow{2}{*}{$L_1$ loss}&\multirow{2}{*}{\makecell[c]{Stepwise Sampling\\Wavelet Transform}}&\multirow{2}{*}{PSNR/SSIM/LPIPS }\\
\cellcolor{gray!20}\cite{Wave-Mamba} &&&&&& \\
\textbf{TSFormer}&&LLIE/Dehaze/Deblur &\multirow{2}{*}{Transformer} &\multirow{2}{*}{$L_1$ loss} &\multirow{2}{*}{\makecell[c]{Stepwise Sampling\\Bilinear, Min-$p$}} &\multirow{2}{*}{PSNR/SSIM/LPIPS}  \\
\cite{tsformer}&&/Derain/Desnow&&&&\\
\shline
\cellcolor{gray!20}\textbf{NSEN}&\multirow{4}{*}{\makecell[c]{Resampling-\\Enhancement}} &\multirow{2}{*}{LLIE/Dehaze} &\multirow{2}{*}{UNet} &\multirow{2}{*}{$L_1$\&Perceptual loss}&\multirow{2}{*}{\makecell[c]{Non-Uniform\\-Sampling}} &\multirow{2}{*}{PSNR/SSIM}\\
\cellcolor{gray!20}\cite{nsen} &&&& &&\\
\textbf{LMAR}&&\multirow{2}{*}{LLIE} &\multirow{2}{*}{MLP} &Smooth$L_1$\&$L_2$&\multirow{2}{*}{\makecell[c]{Model-aware \\Resampling}} &\multirow{2}{*}{PSNR/SSIM/LPIPS}\\
\cite{LMAR} &&&&\&Adversarial
loss &&\\
\shline
\end{tabular}
\end{table*}

\section{Deep Learning-based UHD Image Restoration Approaches}\label{Sec: Deep Learning-based UHD Image Restoration Approaches}
This section presents a comprehensive and systematic review of existing deep learning-based UHD image restoration methods. This review covers various tasks, including super-resolution reconstruction, low-light image enhancement, dehazing, deblurring, deraining, and desnowing. 
%At the same time, we analyze the advantages and disadvantages of representative methods under different tasks.

\subsection{UHD Image Super-Resolution}
Zhang \etal.~\cite{ZhangLLRS0L021} propose the first super-resolution baseline model, MANet, for UHD images. This model consists of four parts: a preprocessing module, a dilation convolution module, a lattice attention module, and an upsampling module. The core idea is to extract features from different depths (horizontal) and different receptive fields (vertical) of the input image, leveraging an attention mechanism to learn the interdependence between feature maps.
This baseline model directly takes low-resolution images as input and is designed for super-resolution tasks.

\subsection{UHD Image Dehazing}
Zheng \etal.~\cite{ZhengRCHWSJ21} originally introduce a multi-guide bilateral learning model for efficient 4K image dehazing on a single GPU, addressing the high computational complexity problem. The model consists of three deep CNNs, and its core principle is to learn the local affine coefficients of bilateral grids from low-resolution images and apply these coefficients to high-resolution features to reconstruct dehazed images.
To further address the challenge of high computational complexity, Xiao \etal.~\cite{LapDehazeNet} develop LapDehazeNet, which integrates the infinite approximation principle of Taylor's theorem with the Laplace pyramid model, for real-time dehazing of 4K images. To mitigate noise introduced by upsampling and downsampling processes, LapDehazeNet introduces a Tucke reconstruction-based regularization term, which is applied to each branch of the pyramid model.
Furthermore, Zhuang \etal.~\cite{dmixer} propose a dimensional transformation mixer (DMixer) specifically designed for UHD industrial camera dehazing. DMixer employs a dual-branch architecture, with high-resolution and low-resolution image features as inputs. Each branch features a dimensional transformation module based on MLPs to capture long-range dependencies within input images.

\subsection{UHD Low-Light Image Enhancement}
To solve the UHD low-light enhancement problem on a single GPU, Lin \etal.~\cite{LinZJ22} follow~\cite{ZhengRCHWSJ21} and also adopt bilateral learning to enhance low-light images. The key difference in obtaining bilateral grids lies in the use of the first-order expansion of Taylor's formula to construct an interpretable network. However, this method is not completely suitable for the high contrast requirements of certain specific tasks.
Subsequently, numerous methods for UHD low-light image enhancement have emerged. For example,
UHDFour~\cite{LiGZLZFL23} enhances the amplitude and phase of low-resolution images in the Fourier domain separately and then adjusts the corresponding high-resolution images at a lower computational cost. This method has demonstrated strong performance on real-world UHD low-light image datasets.
To better model global features, LLFormer~\cite{LLformer} and UHDformer~\cite{uhdformer} both attempt to use Transformers to capture the long-range dependencies of features. However, the attention mechanism of Transformers introduces high computational complexity when processing UHD images. 
Considering the high computational cost and information loss during downsampli
ng in UHD image enhancement tasks, Zou \etal.~\cite{Wave-Mamba} employ the Mamba model for the UHD image enhancement,  combining it with wavelet transform to mitigate information loss. The proposed Wave-Mamba model performs well in low-light image enhancement and noise removal.
Furthermore, several studies have explored the impact of resampling methods on image enhancement. For example, 
to address the limitations of previous downsampling operations that failed to consider content relevance and regional complexity, Yu \etal.~\cite{nsen} propose a non-uniform sampling enhancement network (NSEN) for UHD low-light image enhancement. This model incorporates content-guided downsampling and invertible pixel-alignment. 
The same team also introduces a learning model-aware resampling method (LMAR)~\cite{LMAR}, which fosters collaboration between resizers and enhancers through a custom resampling process guided by model-specific knowledge. This method maintains compatibility with existing interpolative resamplers while significantly enhancing the characterization of low-resolution images.

\subsection{UHD Image Deblurring}
The UHD deblurring method began with UHDVD~\cite{DengRYWSC21}, which introduces a separable-patch architecture combined with a multi-scale integration scheme, specifically designed for UHD video deblurring.
Subsequently, Zheng \etal.~\cite{cub-mixer} propose a multi-scale cubic mixer with a wave-frequency processing framework called Cubic-Mixer. This framework estimates the Fourier coefficients of sharp images from blurry inputs by learning a three-dimensional MLP for real-time UHD image deblurring.

\subsection{UHD Image Deraining $\&$ Desnowing}
Chen \etal.~\cite{UDRMixer} propose an end-to-end trainable multilayer perceptron network, UDR-Mixer, based on a dual-branch architecture for UHD image deraining, achieving favorable performance on both synthetic and real UHD datasets.
In addition to focusing on inherent features, Wang \etal.~\cite{uhddip} incorporate gradient and normal priors in their model design and propose an effective UHD image restoration scheme, UHDDIP, which uses low-resolution prior interactions to guide high-resolution feature reconstruction. The model performs effectively in UHD deraining and desnowing tasks.

\subsection{UHD Image Restoration}
In addition to the above-mentioned solutions for UHD images with single degradation, some researchers are beginning to tackle unified solutions to multiple degradation tasks. For example,
Wang \etal.~\cite{uhdformer} propose a Transformer-based universal UHD image restoration model, called UHDformer. This model modulates reliable low-resolution feature representations to facilitate high-resolution reconstruction by constructing feature transformations from high-resolution space to low-resolution space. UHDformer achieves excellent performance in low-light image enhancement, dehazing, and deblurring with fewer parameters.
Under the dual-branch framework, UHDDIP~\cite{uhddip} incorporates additional prior information in its model design to guide UHD image restoration.
Specifically, UHDDIP utilizes normal and gradient priors to extract useful spatial and detail features in low-resolution space, achieving notable performance across five tasks: low-light image enhancement, dehazing, deblurring, deraining, and desnowing.
%
%UHDPromer introduces implicit neural discriminant prior to measure the difference between high-resolution features and low-resolution features when processing low-resolution features, which is used for UHD low-light image enhancement, dehazing, and deblurring.
%
Moreover, to capture long-range dependencies of features without excessive computational complexity, Yu \etal.~\cite{MixNet} propose an efficient global modeling approach, MixNet, for UHD low-light image enhancement and deblurring.
To balance recovery quality and computational efficiency, Su \etal.~\cite{tsformer} propose the TSFormer, which is capable of processing 4K resolution images in real-time on a single GPU, for five tasks, including low-light enhancement, deblurring, dehazing, deraining, and desnowing. The core principle of the framework is to improve its generalization ability and computational efficiency by integrating trusted learning with sparsified Min-$p$ sampling techniques~\cite{min-p} to generate high-quality attention.
Different from the above methods, Su \etal.~\cite{SimpleIR} propose SimpleIR, an all-in-one UHD image restoration model for four types of degradation: rain, snow, haze, and low light. 
SimpleIR uniformly handles multiple degradations without requiring prior knowledge or prompts and introduces review learning to enhance the restoration model's memory of multiple previously degraded datasets.

\section{Technical Development Review}\label{Sec: Technical Development Review}
In this section, we examine the evolution of deep learning-based UHD image restoration methods, focusing on three critical aspects: network architecture, sampling strategy, and loss function. 

\subsection{Network Architectures}
\noindent Existing UHD image restoration models employ a variety of advanced network architectures, including convolutional neural networks (CNNs)~\cite{KrizhevskySH12}, UNet~\cite{unet2015}, pyramid networks~\cite{Pyramid2017}, multilayer perceptrons (MLPs)~\cite{perceptron1958}, Transformers~\cite{VaswaniSPUJGKP17}, and Mamba~\cite{mamba23}. These architectures provide different processing methods and technical support for UHD image restoration tasks.
To systematically analyze the characteristics and application scenarios of these network structures, we categorize the existing models into the following three forms according to their UHD image processing methods: downsampling-enhancement-upsampling structure, encoder-decoder structure with stepwise up-downsampling, and resampling-enhancement structure.
%%%%%%%%%%%%%%%%%%%%%%%%%%%%%%%

\begin{figure*}[t]
\centering
\begin{center}
\begin{tabular}{c}
\hspace{-2mm}\includegraphics[width=\linewidth]{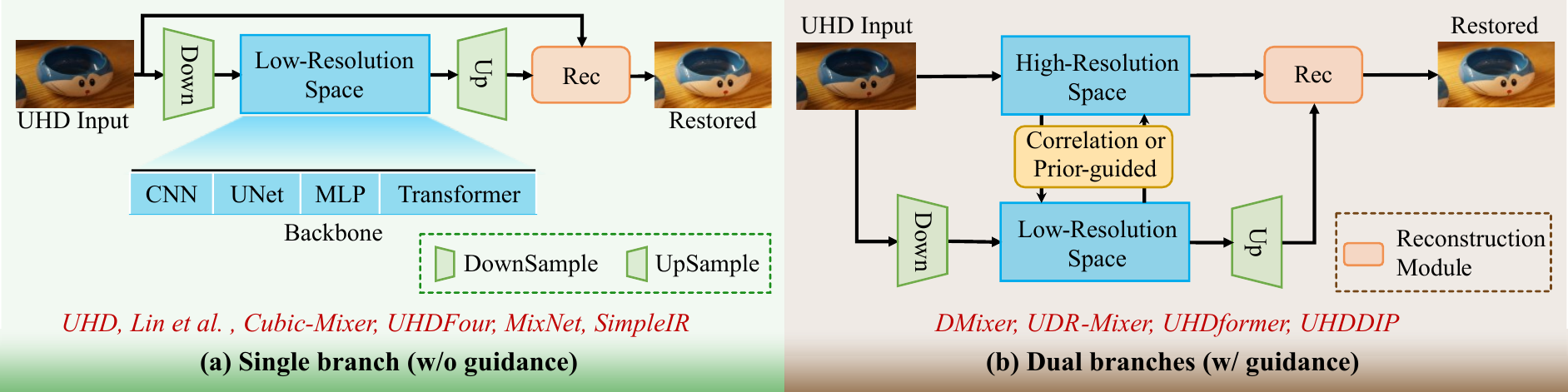} 
\end{tabular}
\caption{\textbf{Summary of the downsampling-enhancement-upsampling structure} for UHD image restoration. (a) The single-branch downsampling-enhancement-upsampling architecture focuses on the design of enhancement networks in the low-resolution space, utilizing some popular architectures such as CNN, UNet, MLP, and Transformer.
(b) The dual-branch downsampling-enhancement-upsampling architecture explores the correlation between high- and low-resolution features or incorporates additional prior information to guide the reconstruction process.
}\label{fig: Downsampling-Enhancement-Upsampling}
\end{center}
\end{figure*}
%%%%%%%%%%%%%%%%%%%
%%%%%%%%%%%%%%%%%%%%%%%%%%%%%%%%%%%%%
\begin{figure*}[t]
\centering
\begin{center}
\begin{tabular}{c}
\hspace{-2mm}\includegraphics[width=\linewidth]{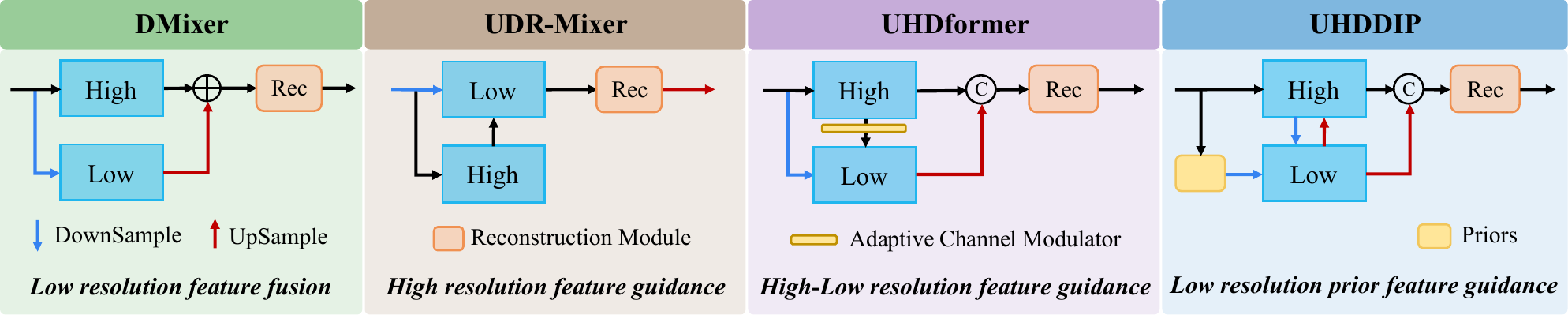} 
\end{tabular}
\caption{Overview of dual branch frameworks under the downsampling-enhancement-upsampling structure.
DMixer~\cite{dmixer} upsamples low-resolution features and merges them with high-resolution features for reconstruction.
UDR-Mixer~\cite{UDRMixer} feeds high-resolution features into the low-resolution branch to facilitate reconstruction.
UHDformer~\cite{uhdformer} transforms features from high to low resolution and enhances high-resolution reconstruction through concatenation.
UHDDIP~\cite{uhddip} extracts gradient and normal priors in the low-resolution space to interact with low-resolution features, guiding high-resolution reconstruction.
}\label{fig: dual-branch}
\end{center}
\end{figure*}
%%%%%%%%%%%%%%%%%%%
\begin{table*}[t]
\setlength{\tabcolsep}{13pt}
\caption{Performance comparison of different guidance mechanisms under the   UHDformer~\cite{uhdformer} framework on the UHD-LL~\cite{LiGZLZFL23} test set.
The best and second-best results are highlighted in \textbf{bold} and \underline{underlined}, respectively.}
\label{tab: different guidance manners} 
\centering
\renewcommand{\arraystretch}{1.3}
\begin{tabular}{l|l|ccc|c}
\shline
\textbf{Framework}
&\textbf{Guidance manner}
&\textbf{~~PSNR~$\uparrow$~} 
&\textbf{SSIM~$\uparrow$~}
&\textbf{LPIPS~$\downarrow$~}
&\textbf{Parameters}
\\
\shline
\multirow{3}{*}{\parbox{2cm}{\textbf{UHDformer}}}
&(a) High resolution feature guidance~\cite{UDRMixer}&25.196 &0.8645 &0.3545 &0.20M \\ 
&(b) Low resolution feature fusion~\cite{dmixer}&\underline{26.750}	&\underline{0.9251}	&\underline{0.2251} &0.34M  \\
&(c) High-Low resolution feature guidance~\cite{uhdformer}&\textbf{27.113} & \textbf{0.9271}&\textbf{0.2240} &0.34M\\  
\shline
\end{tabular}
\end{table*}
%%%%%%%%%%%%%%%%%%%%%%%
%%%%%%%%%%%%%%%%%%%
\begin{figure*}[t]
\centering
\begin{center}
\begin{tabular}{c}
\hspace{-2mm}\includegraphics[width=\linewidth]{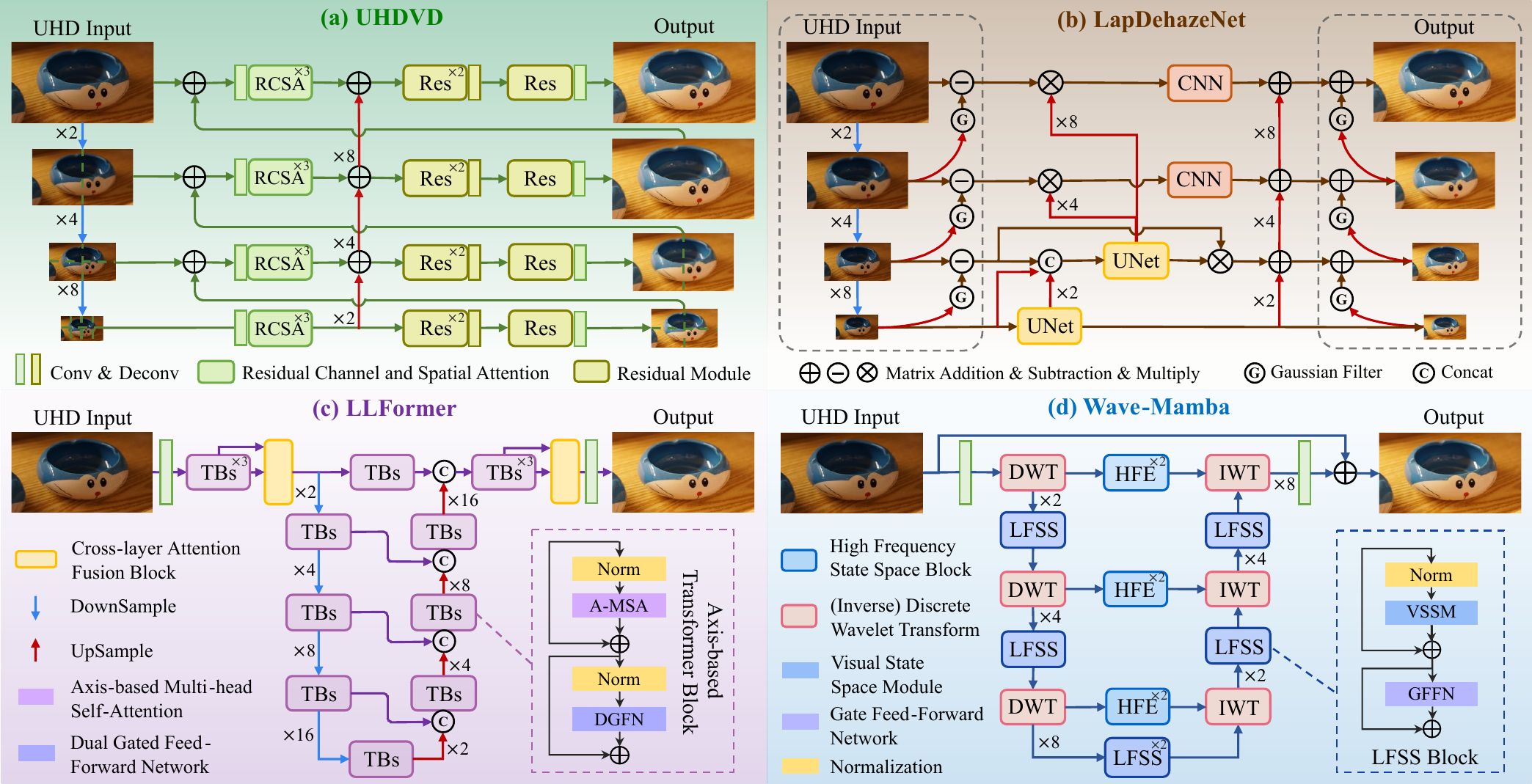}\\
%\hspace{-2mm}\includegraphics[width=\linewidth]{image/Fig6_3.pdf} \\
%\hspace{-2mm}\includegraphics[width=\linewidth]{image/Fig6_3.pdf} 
\end{tabular}
\caption{\textbf{Summary of the Encoder-Decoder structure with stepwise up-downsampling} for UHD image restoration.
UHDVD~\cite{DengRYWSC21} and LapDehazeNet~\cite{LapDehazeNet} progressively downsample the input image based on separable-patch and Laplace pyramid architectures to encode features respectively, while LLFormer~\cite{LLformer} and Wave-Mamba~\cite{Wave-Mamba} reduce the scale of potential features with stepwise and adopt the core components of the Axis-based Transformer block and the Low-Frequency State Space Block (LFSS Block) for restoration, respectively.}\label{fig: Encoder-Decoder with stepwise up-downsampling and resampling structure}
\end{center}
\end{figure*}

\subsubsection{Downsampling-Enhancement-Upsampling}
Early deep learning-based UHD image restoration methods primarily adopted the ``downsampling-enhancement-upsampling" processing paradigm. Specifically, the input UHD image is first downsampled to reduce its resolution. A low-resolution feature representation of the image is learned through an enhancement network, which is then upsampled to a high-resolution space to reconstruct the original UHD image. 
This approach usually uses interpolation and PixelUnshuffe operations for high-magnification upsampling and downsampling to reduce computational burden. Consequently, most research focuses on developing more effective enhancement networks. An illustration of their overall frameworks is shown in Fig.~\ref{fig: Downsampling-Enhancement-Upsampling} and can be further divided into two main categories according to the design of the network branch. The first category focuses on the design of low-resolution enhancement networks using a single branch framework (Fig.~\ref{fig: Downsampling-Enhancement-Upsampling}(a)). 
For example, 
Zhang \etal.~\cite{ZhangLLRS0L021} propose  MANet, a baseline model consisting of convolutional layers, residual blocks, dense blocks, and a new grid attention module for UHD image super-resolution. This attention mechanism enables the network to focus on important features across different depths (horizontal) and levels (vertical) of the receptive field.
Zheng \etal.~\cite{ZhengRCHWSJ21} design a multi-guided bilateral learning framework for UHD image dehazing based on CNN. This framework learns the local affine coefficients of the bilateral grid from low-resolution images and applies them to UHD images. Lin \etal.~\cite{LinZJ22} adopt bilateral learning for UHD low-light enhancement, employing UNet and the first-order unfolding of Taylor's formula to construct an interpretable network.
To achieve real-world UHD low-light image enhancement, Li \etal.~\cite{LiGZLZFL23} propose UHDFour, a CNN-based low-light image enhancement network. UHDFour enhances low-resolution amplitude and phase information in the Fourier domain.
As CNNs focus more on local features, global feature modeling is ignored.
Zheng \etal.~\cite{cub-mixer} propose a multi-scale cubic-mixer based on MLPs for UHD deblurring. This method applies Fourier transform on features at three different scales in a low-resolution space to achieve feature enhancement in the frequency domain.
Wu \etal.~\cite{MixNet} propose MixNet, an efficient model for UHD low-light image enhancement, by modeling global and local feature dependencies using multiple feature mixing blocks (FMBs).
Su \etal.~\cite{SimpleIR} use Transformer to capture long-range dependencies of features, achieving excellent results across multiple tasks, including UHD LLIE, deblurring, deraining, and desnowing.

In addition to single-branch low-resolution enhancement networks, researchers have explored the correlation between high- and low-resolution features in dual-branch frameworks to extract meaningful features for UHD restoration, as shown in Fig.~\ref{fig: Downsampling-Enhancement-Upsampling}(b). The core ideas of these methods are shown in Fig.~\ref{fig: dual-branch}.
Zhuang \etal.~\cite{dmixer} introduce the dimensional transformer mixer model, DMixer, for UHD dehazing. This model is designed as an end-to-end up-and-down two-branch parallel network structure, in which low-resolution branches complement high-resolution branches to enhance the control of global and complementary information at different scales.
Chen \etal.~\cite{UDRMixer} develop UDR-Mixer, consisting of a spatial feature rearrangement block in the low-resolution space and a frequency feature modulation block in high-resolution space. In this framework, the learned high-resolution features assist with low-resolution feature reconstruction to guide image deraining.
Wang \etal.~\cite{uhdformer} propose the UHDformer framework, which incorporates learning in two different spaces. This method provides meaningful feature representations for high-resolution reconstruction by constructing feature transformations from the high-resolution space to the low-resolution space and applying self-attention within the low-resolution space.
To better analyze the impact of different guidance mechanisms on recovery performance, we retrain the model based on the UHDformer~\cite{uhdformer} framework on the low-light dataset UHD-LL~\cite{LiGZLZFL23}, in accordance with the three methods mentioned above. Their test results are summarized in Table~\ref{tab: different guidance manners}.
%A
In addition, leveraging other prior features to guide the modeling of high-resolution features has been explored under a two-branch framework. 
For example,
Wang \etal.~\cite{uhddip} leverage the gradient and normal prior of low-resolution images to design UHDDIP networks containing a prior interaction module that extracts useful spatial and detailed features from the low-resolution space to better guide high-resolution feature recovery.
%
%\textcolor{red}{UHDPromer} incorporates implicit neural discrimination prior, which measures feature differences between high- and low-resolutions to effectively guide the reconstruction of high-resolution features. The network architecture relies primarily on a Transformer framework.

These methods described above mainly focus on the design of enhancement networks, including the direct learning of self-feature representation and exploring the interactions of other prior features to facilitate the reconstruction of high-resolution features. 
%
%Table.~\ref{tab:different backbones} shows a performance comparison for different backbones under the UHDformer~\cite{uhdformer} framework on UHD-LL~\cite{LiGZLZFL23} test set.
%
%\textcolor{red}{It is evident that both UHDformer~\cite{uhdformer} and UHDDIP~\cite{uhddip}, which utilize Transformer-based backbone networks, outperform UHDFour~\cite{LiGZLZFL23}, which relies on a CNN-based backbone. On the other hand, UHDformer achieves the highest PSNR by leveraging its low-resolution features to guide high-resolution recovery, which means it can improve image fidelity. Conversely, UHDDIP excels in SSIM by employing prior information to guide reconstruction, reflecting its strength in preserving image structural integrity.}
%
However, when applied to UHD images, the use of high-magnification up- and down-sampling results in significant information loss, which adversely affects image restoration.
%%%%%%%%%%%%%%%%%%%
\begin{table*}[!th]
\setlength{\tabcolsep}{14pt}
\caption{Performance comparison of different backbones under the UHDformer~\cite{uhdformer} framework on the UHD-LL~\cite{LiGZLZFL23} test set.
The best and second-best results are highlighted in \textbf{bold} and \underline{underlined}, respectively.}
\label{tab:different backbones} 
\centering
\renewcommand{\arraystretch}{1.3}
\begin{tabular}{l|cc|ccc|c}
\shline
\textbf{Framework}
&\textbf{Backbone}
&\textbf{Core block}
&\textbf{~~PSNR~$\uparrow$~} 
&\textbf{SSIM~$\uparrow$~}
&\textbf{LPIPS~$\downarrow$~}
&\textbf{Parameters}
\\
\shline
\multirow{3}{*}{\parbox{2cm}{\textbf{UHDformer}}}
&CNN&ConvNext~\cite{uhdformer}&21.387 &0.9083 &0.2567&0.22M  \\
&Mamba&LFSS block~\cite{Wave-Mamba}&\underline{24.815}&\underline{0.9165}&\underline{0.2332} &0.36M\\  
&Transformer&Transformer Block~\cite{Zamir2021Restormer}&
\textbf{26.767} &\textbf{0.9259}&\textbf{0.2086}
%25.233 	 &0.9195& 0.2130 
&0.26M \\ 
\shline
\end{tabular}
\end{table*}
%%%%%%%%%%%%%%%%%%%%%%%%%%%%%%%%%%%%%%%
\begin{figure*}[t]
\centering
\begin{center}
\begin{tabular}{c}
\hspace{-2mm}\includegraphics[width=\linewidth]{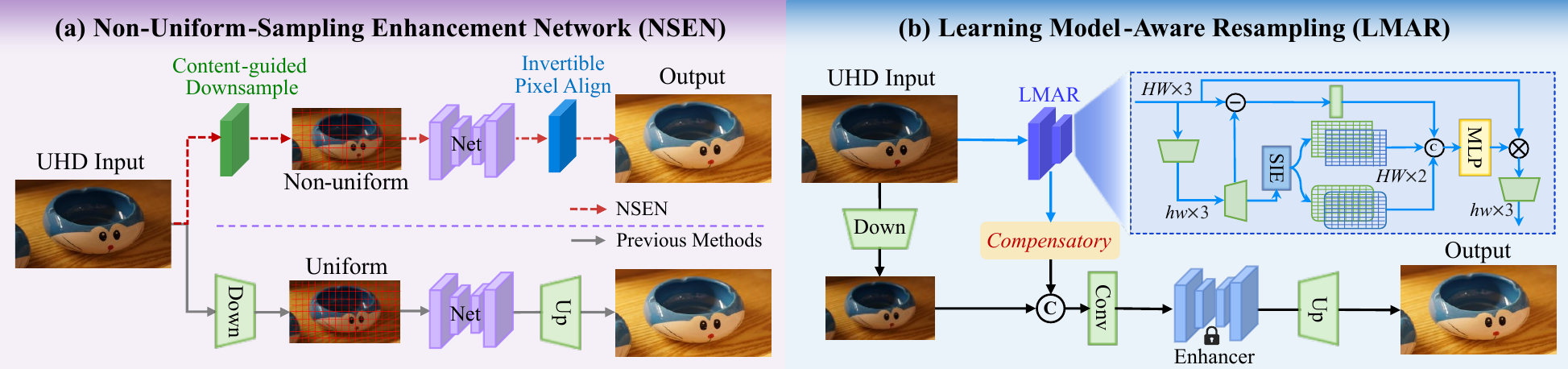} 
\end{tabular}
\caption{\textbf{Summary of the resampling-enhancement structure} for UHD image restoration. Unlike previous methods that rely on uniform, content-agnostic downsampling (represented by the gray arrows), (a) a non-uniform sampling enhancement network (NSEN)~\cite{nsen}, which incorporates two core designs: 1) content-guided downsampling to generate detail-preserving low-resolution images, and 2) invertible pixel alignment that computes inverse functions to remove distortions induced during the downsampling process; and (b) a learning model-aware resampling (LMAR)~\cite{LMAR}, which focuses on obtaining compensated low-resolution features from UHD input images guided by model knowledge. These features are fed into the enhancer, along with original low-resolution features, and are subsequently upsampled to UHD results.
}\label{fig: resampling-enhancement structure}
\end{center}
\end{figure*}
\subsubsection{Encoder-Decoder with Stepwise Up-downsampling}
The encoder-decoder structure with stepwise up-downsampling is another architecture used for UHD image restoration. Its core concept is to extract and process different low-resolution features through stepwise downsampling during the encoder stage and to progressively reconstruct high-resolution features through stepwise upsampling during the decoder stage. This type of structure is summarized in Fig.~\ref{fig: Encoder-Decoder with stepwise up-downsampling and resampling structure}.
Deng \etal.~\cite{DengRYWSC21} propose an asymmetric encoder-decoder structure based on separable patches, named UHDVD, for UHD video deblurring. In the encoder-decoder stage, the residual channel-spatial attention (RCSA) module and the normal residual block are separately employed to extract patch features at different scales, which are then fused. This design enables fast and accurate deblurring without requiring additional computational resources.
Wang \etal.~\cite{LLformer} propose a hierarchical encoder-decoder structure, LLFormer, based on Transformers for UHD low-light image enhancement. Its core innovation lies in the integration of axis-based multi-head self-attention mechanism and the cross-layer attention fusion module to enhance global features at different resolutions. However, applying  Transformers to process UHD images significantly increases computational complexity.
To address this issue, Xiao \etal.~\cite{LapDehazeNet} use the Laplacian pyramid to process multi-resolution features. They designed LapDehazeNet, guided by Taylor's theorem, for UHD image dehazing. This network effectively reduces computational complexity while maintaining dehazing performance.
Su \etal.~\cite{tsformer} introduce TSFormer, an encoder-decoder architecture composed of multiple hierarchical levels of trusted sparse blocks (TSBs), as a cost-effective alternative to the multi-head self-attention module. In the encoder, a sequence of TSBs, focused on trusted mechanisms and Min-$p$ sampling, progressively downsamples the feature map. The decoder then reconstructs the high-resolution in a progressive manner.
In addition, Zou \etal.~\cite{Wave-Mamba} replaced Transformer with the more computationally efficient Mamba model and proposed the Wave-Mamba model, which combines wavelet transform for UHD low-light image enhancement.
Table~\ref{tab:different backbones} presents a performance comparison of various backbones under the UHDformer~\cite{uhdformer} framework on the UHD-LL~\cite{LiGZLZFL23} test set.
Specifically, ConvNext, Transformer blocks, and LFSS blocks are utilized as the core components of the two branches and reconstruction modules, respectively.
The results demonstrate that the Transformer-based model achieves the best performance, followed by the Mamba-based model. Although the LFSS block improves performance compared to the CNN-based network, it also leads to an increase in the number of parameters.

%It can be seen that when the Transformer serves as the backbone, TSFormer~\cite{tsformer} obtains the best PSNR and SSIM scores by introducing the Min-$p$ sampling technique, which also implies that downsampling like Pixelshuffle and interpolation operations will have difficulty in reconstructing the lost information, whereas reliable features will be retained with the addition of Min-$p$ sampling to improve the image recovery quality. Wave-Mamba~\cite{Wave-Mamba}, although slightly inferior in performance, has the smallest number of parameters due to its linear computational complexity. Furthermore, the wavelet transform circumvents the information loss typically associated with downsampling, providing a notable performance advantage over LLformer~\cite{LLformer}.

\subsubsection{Resampling-Enhancement}\;
%The latest research explores the impact of resampling operations on the UHD image features restored by the enhancement network.
%
To preserve the richness of detail and content in low-resolution images after downsampling, several approaches have investigated the impact of resampling operations on the feature representation capabilities of enhancement network. This process is illustrated in Fig.~\ref{fig: resampling-enhancement structure}.
Yu \etal.~\cite{nsen} propose a non-uniform sampling enhancement network (NSEN) for UHD low-light image enhancement, consisting of two key modules: content-guided downsampling and invertible pixel alignment. 
The content-guided downsampling module uses a spatial-variant and invertible non-uniform downsampler to adaptively adjust the sampling rate based on the richness of image details, while the invertible pixel alignment module iteratively remaps the forward sampling process.
Furthermore, they introduce Learning Model-Aware Resampling (LMAR)~\cite{LMAR}, which establishes an association between the sampling algorithm and the enhancement algorithm by customizing resampling guided by model knowledge. 
Importantly, this approach is fully compatible with existing interpolation resamplers, without requiring retraining of the enhancement network, effectively enhancing UHD low-light images. 
This framework ensures consistency between features of UHD image inputs and low-resolution inputs while preserving the representational capacity of intermediate features.
%This structure ensures the consistency of features between UHD image input and low-resolution input, as shown in Fig.~\ref{fig: Encoder-Decoder with stepwise up-downsampling and resampling structure}(d).
%%%%%%%%%%%%%%%%
\begin{table}[b]
\setlength{\tabcolsep}{3.25pt}
\caption{Performance comparison of different sampling methods on a factor of $8\times$ under the UHDformer~\cite{uhdformer} framework on the UHD-LL~\cite{LiGZLZFL23} test set.
The best and second-best results are marked in \textbf{bold} and \underline{underlined}, respectively.}
\label{tab:different sampling strategy} 
\centering
\renewcommand{\arraystretch}{1.3}
\begin{tabular}{l|c|ccc}
\shline
\textbf{Framework}
&\textbf{Sampling Methods}
&\textbf{~~PSNR~$\uparrow$~} 
&\textbf{SSIM~$\uparrow$~}
&\textbf{LPIPS~$\downarrow$~}
%&\textbf{Parameters}
\\
\shline
\multirow{3}{*}{\textbf{UHDformer}}
&Bicubic interpolation &26.672 &0.9259 &0.2292 \\%&0.17M    \\
&Bilinear interpolation&\underline{27.085} &\underline{0.9268} &\textbf{0.2234}\\ %&0.17M  \\
&Pixelshuffle &\textbf{27.113} & \textbf{0.9271} &\underline{0.2240} \\ %&0.34M    \\
\shline
\end{tabular}
\end{table}
%%%%%%%%%%%%%%%%%%%
\subsection{Sampling Strategy}
In existing UHD image restoration frameworks, three primary sampling strategies are commonly used: (1) high magnification up-downsampling (factors with of 4$\times$, 8$\times$, 16$\times$ and 32$\times$), (2) stepwise up-down sampling (factors decreasing from 2$\times$ to 16$\times$), and (3) content-related resampling.

As shown in Table~\ref{tab:Deep Learning-Based UHD Image Restoration Methods}, most methods utilize bicubic interpolation~\cite{LinZJ22,ZhengRCHWSJ21,cub-mixer,dmixer}, bilinear interpolation~\cite{DengRYWSC21,LapDehazeNet}, and PixelUnshuffle~\cite{uhdformer,LiGZLZFL23,uhddip,MixNet,UDRMixer,SimpleIR} to downsample UHD images at high magnifications. This approach is preferred because directly processing UHD images introduces substantial computational challenges. 
We conduct a performance comparison of different sampling
methods with a factor of $8\times$ including bicubic, bilinear interpolation, and pixelshuffle, using the  UHDformer framework on the UHD-LL~\cite{LiGZLZFL23} test set, as shown in Table~\ref{tab:different sampling strategy}. The results demonstrate that the pixelshuffle operation achieves the best performance in terms of PSNR and SSIM, while bilinear interpolation performs best in terms of LPIPS.
However, these methods under high up-downsampling magnification lead to significant information loss, ultimately degrading image quality during restoration.

On the other hand, stepwise up-downsampling~\cite{LapDehazeNet,LLformer} is often integrated into encoder-decoder structures to extract multi-scale features from UHD input images. Despite its advantages, as the degree of downsampling increases, the resulting damage to image quality becomes irreversible during the subsequent upsampling process. 
To address this issue, the wavelet transform is employed for up-downsampling to avoid the loss of critical information~\cite{Wave-Mamba}.
Su \etal.~\cite{tsformer} also introduce Min-$p$ sampling for UHD image restoration, aiming to retain high-confidence features while discarding less important features for effective sparsification.

Additionally, to consider content relevance, Yu \etal.~\cite{nsen} propose a spatial-variant and invertible non-uniform downsampler that adaptively adjusts the sampling rate according to the richness of image details.
LMAR~\cite{LMAR} adopts a model-aware resampling strategy, which uses model-driven knowledge to customize resampling. This approach ensures compatibility with existing interpolative resamplers while eliminating the need to retrain enhancement networks.

\subsection{Loss Functions}
From Table~\ref{tab:Deep Learning-Based UHD Image Restoration Methods}, various loss functions are used to supervise model training for UHD image restoration, including pixel-level loss, SSIM loss, frequency-domain loss, perceptual loss, and adversarial loss.
This section provides a detailed explanation of several loss functions commonly used in UHD image restoration methods.
\subsubsection{Pixel-Level Loss}
The pixel-level loss measures the difference between the recovered image and the clean image in pixel space. The most commonly used loss functions are the $L_1$ loss (Mean Squared Error, MAE) and $L_2$ loss (Mean Squared Error, MSE), defined as follows:
\begin{equation}
\begin{array}{ll}
L_1=||y-\hat{y}||,
\end{array}
\label{eq:l1loss}
\end{equation}
\begin{equation}
\begin{array}{ll}
L_2=||y-\hat{y}||^2,
\end{array}
\label{eq:l2loss}
\end{equation}
where $y$ and $\hat{y}$ are the ground truth and the recovered image, respectively. The quality of the restored image is measured by calculating the absolute difference or the pixel-squared difference between the two images. $L_2$ loss tends to penalize larger errors more significantly and smooth the restored result. $L_1$ loss is more robust to outliers and better at retaining image details. Therefore, most UHD image restoration~\cite{cub-mixer,dmixer,uhdformer,LiGZLZFL23,MixNet,SimpleIR,LinZJ22,uhddip,Wave-Mamba,UDRMixer} efforts rely on $L_1$ loss, while some methods~\cite{ZhengRCHWSJ21,DengRYWSC21,LMAR} use $L_2$ loss.

$Smooth L_1$ loss~\cite{Girshick15} combines the advantages of both $L_1$ loss and $L_2$ loss by using a piecewise function to reduce sensitivity to outliers. It is defined as follows:
\begin{equation}
\begin{array}{ll}
L_{smoothl_1}=\begin{cases}0.5||y-\hat{y}||^2& \text{if $||y-\hat{y}||<1$}\\
||y-\hat{y}||-0.5& \text{otherwise}
\end{cases},
\end{array}
\label{eq:smoothl1loss}
\end{equation}

$Charbonnier$ loss is a smoothed $L_2$ loss, defined as follows:
\begin{equation}
\begin{array}{ll}
L_{char}=\sqrt{||y-\hat{y}||^2+\epsilon^2},
\end{array}
\label{eq:Charbonnierloss}
\end{equation}
where $\epsilon$ is a constant introduced to avoid the error being zero.

$TV$ (Total Variation) loss improves image smoothness and reduces noise and artifacts~\cite{DengRYWSC21} by constraining gradient changes.

While these above loss functions focus on pixel-level consistency, which may lead to over-smoothing and fail to adequately capture perceptual quality.
\subsubsection{SSIM Loss}
$SSIM$ loss~\cite{SSIM_wang} measures the similarity between the restored image and the clean image in terms of brightness, contrast, and structure, defined as follows:
\begin{equation}
\begin{array}{ll}
L_{ssim}=\dfrac{(2\mu_y\mu_{\hat{y}}+c_1)(2\sigma_{y\hat{y}}+c_2)}{(\mu_y^2+\mu_{\hat{y}}^2+c_1)(\sigma_y^2+\sigma_{\hat{y}}^2+c_2)},
\end{array}
\label{eq:SSIMloss}
\end{equation}
where $\mu$, $\sigma$, and $\sigma_{y\hat{y}}$ represent the mean, standard deviation, and covariance respectively. $c_1$ and $c_2$ are constants, introduced to ensure numerical stability.

\subsubsection{Frequency-Domain Loss}
The frequency loss~\cite{JiangDWL21} evaluates the error between the predicted image and the clean image in the frequency domain, expressed as follows:
\begin{equation}
\begin{array}{ll}
L_{freq}=||\mathcal{F}(y)-\mathcal{F}(\hat{y})||_1,
\end{array}
\label{eq:frequencyloss}
\end{equation}
where $\mathcal{F}$ denotes the Fast Fourier transform. Due to its emphasis on high-frequency details, studies in~\cite{uhdformer,uhddip,SimpleIR} employ a weighted sum of frequency loss and other pixel-level losses for UHD image restoration.

In addition, Fourier spatial loss, as proposed by~\cite{LinZJ22} for UHD low-light image enhancement, calculates the amplitude and phase angle differences between the output image and the ground-truth image. By performing weighted averaging, it improves the restoration of missing high-frequency content in the input image.

\subsubsection{Perceptual Loss}
Perceptual loss~\cite{JohnsonAF16} focuses on differences between a restored image and its ground-truth image in the feature space of a pretrained network (such as VGG), aligning more closely with human visual perception. It emphasizes high-level structure and semantic information over pixel-level accuracy. The loss is defined as follows:
\begin{equation}
\begin{array}{ll}
L_{perc}=\sum_{l}||\phi_l(y)-\phi_l(\hat{y})||_1,
\end{array}
\label{eq:perceptualloss}
\end{equation}
where $\phi_l$ represents the feature mapping of the $l$-th layer of the pretrained network.
%%%%%%%%%%%%%%%%%%%
\begin{table}[b]
\setlength{\tabcolsep}{8.5pt}
\caption{Performance comparison on the UHD-LL~\cite{LiGZLZFL23} test set using several common loss functions under the UHDformer~\cite{uhdformer} framework.
The best and second-best results are marked in \textbf{bold} and \underline{underlined}, respectively.}
\label{tab:Performance comparison of different loss function.} 
\centering
\renewcommand{\arraystretch}{1.3}
\begin{tabular}{l|ccc}
\shline
\textbf{Loss Function}
&\textbf{~~PSNR~$\uparrow$~} 
&\textbf{SSIM~$\uparrow$~}
&\textbf{LPIPS~$\downarrow$~}
\\
\shline
$L_1$ Loss&24.827  &0.8826 &0.3551\\ 
$L_2$ Loss&24.905 &0.8843 &0.3502\\ 
SSIM Loss&24.408 &0.8725 &0.3929\\
Frequency Loss &\underline{26.751} &\textbf{0.9296} &\textbf{0.2185}\\ 
$L_1$ + SSIM Loss & 23.597 &0.8765 &0.3635\\ 
$L_2$ + SSIM Loss & 25.900
&0.8898 &0.3089\\ 
$L_1$ + Frequency Loss &\textbf{27.113} &0.9271 &0.2240\\ 
$L_2$ + Frequency Loss &23.370 & 0.8918 &0.3278\\ 
SSIM + Frequency Loss &26.738 & \underline{0.9275} &\underline{0.2216} \\

\shline
\end{tabular}
\end{table}
%%%%%%%%%%%%%%%%%%%
\subsubsection{Adversarial Loss}
Adversarial loss~\cite{GoodfellowPMXWOCB14} originates from the Generative Adversarial Networks (GANs). It determines the difference between the generated image $y$ and the real image $\hat{y}$ using a discriminator $D$, expressed as follows:
\begin{equation}
\begin{array}{ll}
L_{adv}=logD(y)+log(1-D(\hat{y})),
\end{array}
\label{eq:adversarialloss}
\end{equation}

LMAR~\cite{LMAR} adopts adversarial loss to promote the interaction between the resizer and the enhancer, ensuring consistency in the distribution consistency of original downsampled features and resampled features.

To evaluate the impact of various loss functions on the UHD image restoration task, we conduct experiments on the UHD-LL~\cite{LiGZLZFL23} test set under the UHDformer~\cite{uhdformer} framework.
With the same experimental configurations and the number of iterations, the results are summarized in Table~\ref{tab:Performance comparison of different loss function.}.
The findings reveal that using frequency loss alone yields the best SSIM and LPIPS scores, demonstrating its effectiveness in preserving image structure and perceptual quality. However, its relatively low PSNR score indicates limitations in enhancing image fidelity.
On the other hand, $L_1$ and $L_2$ losses excel at improving image fidelity. Combining $L_1$ loss with frequency loss, denoted $L_1$ $+$ frequency loss, achieves the highest PSNR score while maintaining third-best performance in terms of both SSIM and LPIPS, striking a balance between fidelity and perceptual quality.

\section{Performance Evaluation}\label{Sec: Performance Evaluation}
In this section, we first introduce the evaluation metrics used for UHD image restoration, followed by a performance comparison of several representative UHD image restoration methods and a complexity analysis of the main algorithms.
\subsection{Evaluation Metrics}
In image quality assessment, evaluation metrics are categorized into two main types: full-reference metrics and no-reference metrics. These metrics are used to evaluate the quality of restored images by either comparing them to ground-truth images or assessing them without reference images.
For the experiments in Sec.~\ref{sec:Comparision Results}, we utilize the commonly used IQA PyTorch Toolbox~\footnote{ https://github.com/chaofengc/iqa-pytorch} to compute popular metric scores.
\subsubsection{Full-Reference Metrics}
\noindent\textbf{PSNR.}\; Peak Signal-to-Noise Ratio (PSNR)~\cite{PSNR_thu} measures the pixel difference between the reconstructed image and the original or reference image. It is widely used in restoration tasks,  with higher PSNR values indicating better image reconstruction quality. However, as it is based solely on pixel differences, PSNR may not accurately reflect human visual perception of image quality.

\noindent\textbf{SSIM.}\; Structural Similarity Index Measure (SSIM)~\cite{SSIM_wang} evaluates image quality by comparing the structural information of the original and restored images.
It accounts for brightness, contrast, and structural differences, making it sensitive to human visual perception. Higher SSIM values indicate better quality.

\noindent\textbf{MAE/MSE.}\; Mean Absolute Error (MAE) and Mean Squared Error (MSE) are fundamental metrics used to evaluate the average error between the reconstructed image and its original counterpart. MAE computes the average of the absolute differences between pixel values, whereas MSE measures the average of the squared differences, placing greater emphasis on larger errors due to its quadratic nature. Similar to PSNR, these metrics focus solely on pixel-wise differences and may not align well with human perception of visual quality.

\noindent\textbf{LPIPS.}\; LPIPS~\cite{Zhang_lpips} assesses perceptual similarity by calculating the distance between feature representations in a neural network.
This deep learning-based metric approximates human judgment of image quality. Lower LPIPS values correspond to higher perceptual similarity.
\begin{table*}[t]
\setlength{\tabcolsep}{9.5pt}
\caption{Low-light image enhancement results on the UHD-LL~\cite{LiGZLZFL23} dataset. The \colorbox{pink!50}{\textbf{best}} and \colorbox{blue!10}{second-best} results for each metric are highlighted.
%The best and second best are marked in \textbf{bold} and \underline{underlined}, respectively.
$\uparrow$($\downarrow$) means that higher(lower) values are better.}
\label{tab:Low-light image enhancement_UHDLL.} 
\centering
\renewcommand{\arraystretch}{1.3}
\begin{tabular}{l|l|c|ccc|ccc}
\shline
\textbf{Type}
&\textbf{Methods}
&\textbf{Venue}
&\textbf{~~PSNR~$\uparrow$~} 
&\textbf{SSIM~$\uparrow$~}
&\textbf{LPIPS~$\downarrow$~}
&\textbf{~~NIQE~$\downarrow$~} 
&\textbf{MUSIQ~$\uparrow$~}
&\textbf{PI~$\downarrow$~}
\\
\shline
\multirow{3}{*}{\textbf{Non-UHD}} 
&SwinIR~\cite{liang2021swinir}& ICCVW’21&21.165 & 0.8450 &0.3995&8.3811	&21.6384	&7.2767 \\ 
&Restormer~\cite{Zamir2021Restormer}&CVPR’22&21.536 & 0.8437 &0.3608&8.2193	&23.2947	&7.0071 \\ &Uformer~\cite{wang2021uformer}&CVPR’22&21.303   & 0.8233  &0.4013 & 8.8050	&21.8972	&6.9021 \\
\shline
\multirow{5}{*}{\textbf{UHD}} 
&LLFormer~\cite{LLformer}&AAAI’23&24.065 & 0.8580 &0.3516 & 8.2977&	21.1605	&7.4717 \\
&UHDFour~\cite{LiGZLZFL23}&ICLR’23&26.226 & 0.9000 &0.2394  &\cellcolor{pink!50}{\textbf{6.0459}}	&28.5186	&\cellcolor{pink!50}{\textbf{6.2463}} \\
&UHDformer~\cite{wang2024uhdformer}&AAAI’24&\cellcolor{blue!10}{27.113} & \cellcolor{blue!10}{0.9271} &0.2240 &7.0614&	35.8419&	6.5761   \\ 
&UHDDIP~\cite{uhddip}&ArXiv’24&26.749&\cellcolor{pink!50}{\textbf{0.9281}}&\cellcolor{blue!10}{0.2076} &6.9962	& \cellcolor{blue!10}{37.5337}	&6.5356  \\
%&UHDPromer&-&\underline{27.159}&\textbf{0.928} &0.211  &7.222&	\underline{37.279}&	6.674 \\
&Wave-Mamba~\cite{Wave-Mamba}&ACM MM’24&\cellcolor{pink!50}{\textbf{27.352}} & 0.9123 &\cellcolor{pink!50}{\textbf{0.1853}} &\cellcolor{blue!10}{6.9379} &\cellcolor{pink!50}{\textbf{41.9605}} &\cellcolor{blue!10}{6.4766}	  \\ 
%&TSFormer~\cite{tsformer}&ArXiv’24&\textbf{27.40} & \textbf{0.933} &- &- &- &-	  \\ 
\shline
\end{tabular}
\end{table*}
%%%%%%%%%%%%%%%%%%
\begin{figure*}[t]
\centering
\begin{center}
\begin{tabular}{c}
\hspace{-2mm}\includegraphics[width=1\linewidth]{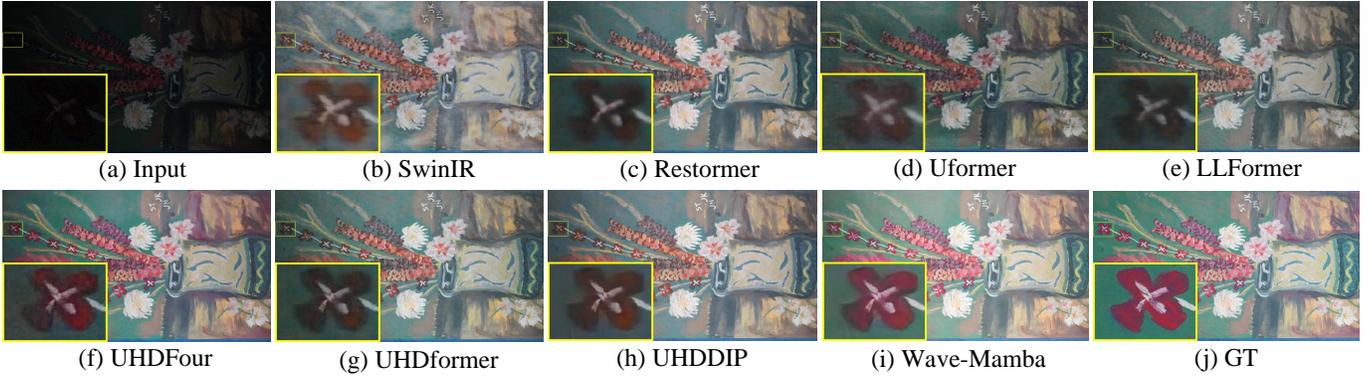}\\
\end{tabular}
\caption{Visual results of different methods for UHD low-light image enhancement on the UHD-LL~\cite{LiGZLZFL23} dataset.}
\label{fig:LLIE on UHD-LL}
\end{center}
\end{figure*}
%%%%%%%%%%%%%%%%%%%%%%%%%
\subsubsection{No-Reference Metrics}

\noindent\textbf{NIQE.}\; The primary purpose of Natural Image Quality Evaluator (NIQE)~\cite{niqe_2013} is to assess an image's naturalness based on statistical models of natural scenes. It determines how visually natural and realistic an image appears without requiring reference images. Lower NIQE scores indicate more natural and realistic images.

\noindent\textbf{MUSIQ.}\; Multi-scale Image Quality (MUSIQ)~\cite{musiq} evaluates image quality by analyzing contrast preservation across multiple scales. It focuses on the preservation of fine details and textures in the image after processing. Higher MUSIQ values signify better image quality.

\noindent\textbf{PI.}\; Perceptual Index (PI) combines two metrics, MAE and NIQE, to evaluate image perceptual quality. It emphasizes both aesthetic appeal and naturalness. A lower PI score reflects better perceptual quality.

\subsection{Comparision Results}\label{sec:Comparision Results}
In this section, we evaluate the quantitative and qualitative performance of representative UHD image restoration methods across a range of tasks, including low-light image enhancement, image dehazing, deblurring, deraining, and desnowing.
To ensure a comprehensive comparison, we present test results using the published pre-trained models of these UHD methods. Additionally, we include quantitative results from published studies for other methods where pre-trained models are not available. 

It is worth noting that some non-UHD methods are not able to directly process UHD images during evaluation. To address this limitation, two approaches are typically employed. The first approach resizes the input image to the maximum size that the model supports, then restores it to its original resolution after processing~\cite{LiGZLZFL23,uhdformer,uhddip}. While straightforward, this method may result in information loss.
The second approach divides the input UHD image into smaller patches, processes them independently, and stitches them together to form the final restored image~\cite{Wave-Mamba}. While this patch-based strategy accommodates larger images, it may introduce artifacts along the patch boundaries.

\subsubsection{UHD Low-Light Image Enhancement}
%
% In this section, we compare the performance of various advanced UHD low-light image enhancement (LLIE) methods using two widely adopted UHD LLIE datasets: UHD-LL~\cite{LiGZLZFL23} and UHD-LOL4K~\cite{LLformer}. UHD-LL contains 150 pairs of real-world test images, while UHD-LOL4K includes 2,100 pairs of synthesized low-light/normal-light images.
In this section, we compare the performance of various advanced UHD low-light image enhancement (LLIE) methods on the widely used UHD-LL~\cite{LiGZLZFL23} dataset, which contains 150 pairs of real-world test images.
\\
\\
\noindent\textbf{Quantitative Comparisons:}
For the evaluation on the UHD-LL dataset, we present the results of three non-UHD methods (SwinIR~\cite{liang2021swinir}, Restormer~\cite{Zamir2021Restormer}, and Uformer~\cite{wang2021uformer}), and five UHD LLIE methods: LLFormer~\cite{LLformer}, UHDFour~\cite{LiGZLZFL23}, UHDFormer~\cite{uhdformer}, UHDDIP~\cite{uhddip}, and Wave-Mamba~\cite{Wave-Mamba}. The results of the non-UHD methods are obtained from~\cite{uhdformer}, while those for the UHD methods are generated using code or pre-trained models published by the respective authors. The final performance results are summarized in Table~\ref{tab:Low-light image enhancement_UHDLL.}.

To comprehensively evaluate performance, in addition to full-reference metrics (PSNR, SSIM, and LPIPS), we also compute three no-reference metrics: NIQE, MUSIQ, and PI, to provide additional insights into perceptual quality and overall aesthetic appeal.
The results reveal that UHD methods significantly outperform non-UHD methods.
Among the UHD methods, UHDDIP~\cite{uhddip} achieves the highest SSIM score, indicating the effectiveness of using additional prior information to preserve image texture and structural details. Conversely, Wave-Mamba~\cite{Wave-Mamba} delivers the best performance in terms of PSNR, LPIPS, and MUSIQ, excelling in perceptual quality, due to the information preservation capability of the wavelet transform.
Moreover, UHDFour~\cite{LiGZLZFL23} performs well in terms of NIQE and PI, reflecting its advantages in restoring the naturalness and realism of images.
% In Table~\ref{tab:Low-light image enhancement_UHD-LOL4K.}, we compare the PSNR, SSIM, and LPIPS metrics of different LLIE methods on the UHD-LOL4K dataset, and these results are obtained from~\cite{Wave-Mamba}.
% %
% It can be observed that MixNet~\cite{MixNet} shows better performance than Wave-Mamba~\cite{Wave-Mamba} due to its efficient modeling on global features.
\\
\\
\noindent\textbf{Qualitative Comparisons:}
Fig.~\ref{fig:LLIE on UHD-LL} illustrates the qualitative performance of various LLIE methods on the UHD-LL test set.
Among the non-UHD methods, SwinIR~\cite{liang2021swinir} demonstrates superior enhancement quality compared to Restormer~\cite{Zamir2021Restormer} and Uformer~\cite{wang2021uformer}. However, it introduces noticeable blurring artifacts, which detract from the overall clarity of the output.
For the UHD methods, both UHDFour~\cite{LiGZLZFL23} and UHDDIP~\cite{uhddip} show promising results in color recovery. However, UHDFour introduces significant noise during the enhancement process, while UHDDIP suffers from edge ghosting artifacts.
In contrast, Wave-Mamba~\cite{Wave-Mamba} achieves a balanced performance by excelling in both color enhancement and detail preservation, making it the most effective approach among the evaluated methods.
\begin{table*}[t]
\setlength{\tabcolsep}{9.7pt}
\caption{Image dehazing results on the UHD-Haze~\cite{uhdformer} dataset. The \colorbox{pink!50}{\textbf{best}} and \colorbox{blue!10}{second-best} results for each metric are highlighted.
%We highlight the \colorbox{pink!50}{\textbf{best}} and \colorbox{blue!10}{second best} values for each metric.
}
\label{tab:image dehazing.} 
\centering
\renewcommand{\arraystretch}{1.3}
\begin{tabular}{l|l|c|ccc|ccc}
\shline
\textbf{Type}
&\textbf{Methods}
&\textbf{Venue}
&\textbf{~~PSNR~$\uparrow$~} 
&\textbf{SSIM~$\uparrow$~}
&\textbf{LPIPS~$\downarrow$~}
&\textbf{~~NIQE~$\downarrow$~} 
&\textbf{MUSIQ~$\uparrow$~}
&\textbf{PI~$\downarrow$~}
\\
\shline
\multirow{3}{*}{\textbf{Non-UHD}} 
&Restormer~\cite{Zamir2021Restormer}&CVPR’22&12.718  &0.6930  &0.4560 & 6.7613 &29.2544 &6.1032  \\
&Uformer~\cite{wang2021uformer}&CVPR’22&19.828 &0.7374 &0.4220  &6.6796	&26.6380	&5.8680 \\
&DehazeFormer~\cite{DehazeFormer}&TIP’23&15.372 &0.7045 &0.3998  & 7.1774	& 29.5049	&6.5477 \\
\shline
\multirow{3}{*}{\textbf{UHD}} 
&UHD~\cite{ZhengRCHWSJ21}&ICCV’21&18.048 &0.8113 &0.3593   &\colorbox{pink!50}{\textbf{4.9918}}&	27.9019&	\colorbox{pink!50}{\textbf{5.0621}} \\
&UHDformer~\cite{wang2024uhdformer}&AAAI’24& \colorbox{blue!10}{22.586} &\colorbox{blue!10}{0.9427} &\colorbox{blue!10}{0.1188} & \colorbox{blue!10}{5.3911}&	\colorbox{blue!10}{31.7302}&	\colorbox{blue!10}{5.2078} \\
%&\textbf{UHDPromer}&-&22.725&	0.943&	0.113 &5.468&	32.742&	5.234\\
&UHDDIP~\cite{uhddip}&ArXiv’24&\colorbox{pink!50}{\textbf{24.699}} &\colorbox{pink!50}{\textbf{0.9520}} &\colorbox{pink!50}{\textbf{0.1049}} & 5.4714&	\colorbox{pink!50}{\textbf{33.0505}}&	5.2490 \\
%&TSFormer~\cite{tsformer}&ArXiv’24&\textbf{24.88} & \textbf{0.953} &\textbf{0.092} &-&-&- \\ 
\shline
\end{tabular}
\end{table*}
%%%%%%%%%%%%%%%%%%%%%%
\begin{figure*}[!th]
\centering
\begin{center}
\begin{tabular}{c}
\hspace{-2mm}\includegraphics[width=1\linewidth]{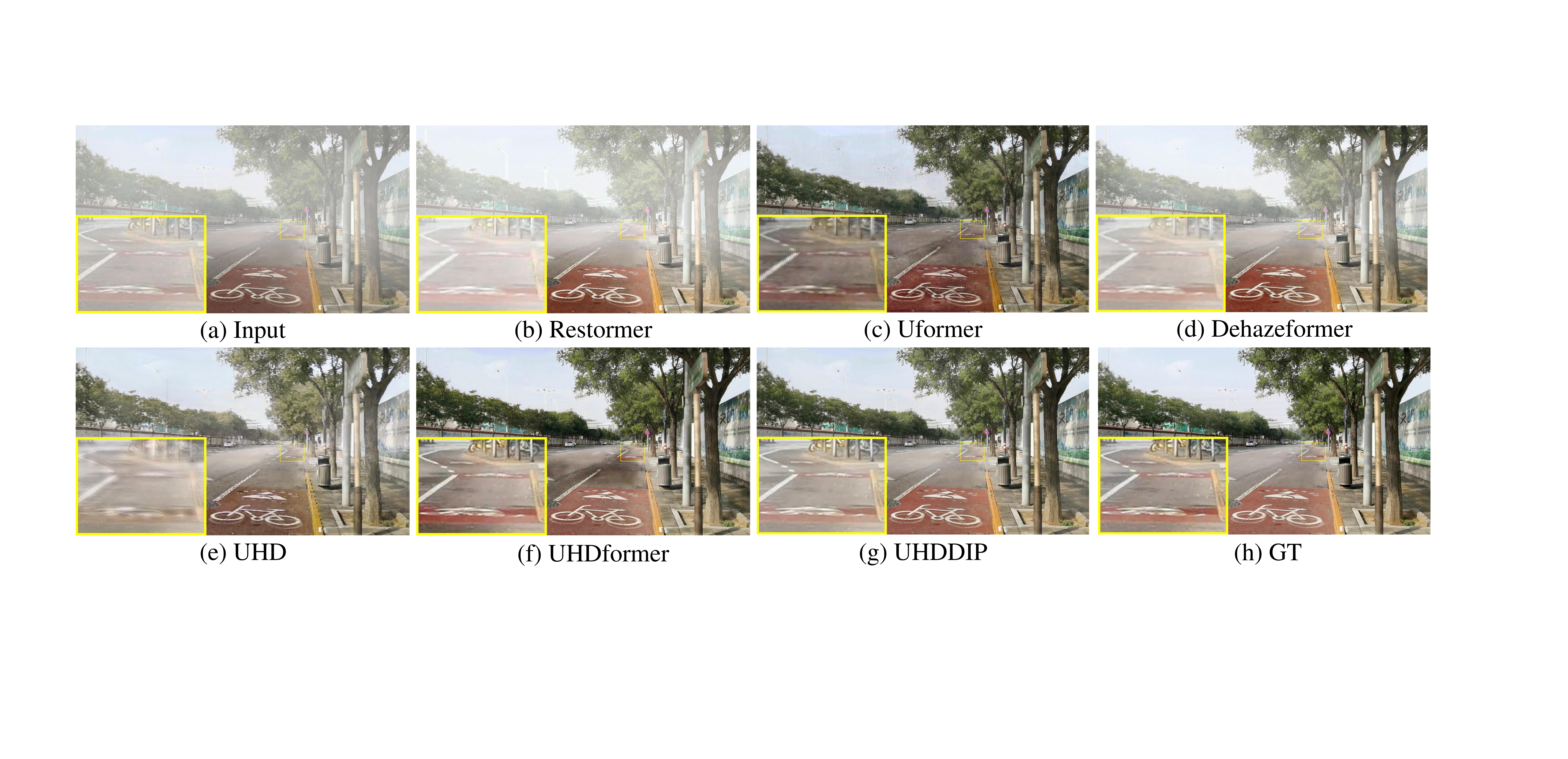}
\\
\end{tabular}
\caption{Visual results of different methods for UHD image dehazing on the UHD-Haze~\cite{uhdformer} dataset.}
\label{fig:image dehazing on UHD-Haze}
\end{center}
\end{figure*}
%%%%%%%%%%%%%%%%%%%%%%%%%

\subsubsection{UHD Image Dehazing}
The comparison of various methods tested on the 4KID dataset~\cite{ZhengRCHWSJ21} is limited by inconsistencies in the number of datasets used.
In this section, we evaluate six methods on the UHD-Haze dataset~\cite{uhdformer}, which contains 230 paired images.  These include three non-UHD methods~\cite{Zamir2021Restormer,wang2021uformer,DehazeFormer}, and three representative UHD image dehazing models: UHD~\cite{ZhengRCHWSJ21}, UHDformer~\cite{uhdformer}, and UHDDIP~\cite{uhddip}. 
\\
\\
\noindent\textbf{Quantitative Comparisons:}
In Table~\ref{tab:image dehazing.}, we evaluate the recovery performance of different dehazing methods using six widely used metrics.
Leveraging additional prior information, UHDDIP~\cite{uhddip} shows excellent performance across several full-reference metrics (PSNR, SSIM, and LPIPS), as well as the no-reference metric MUSIQ.
In contrast, UHD~\cite{ZhengRCHWSJ21}, which employs bilateral learning, performs well on the no-reference metrics NIQE and PI.
In addition, UHDformer~\cite{uhdformer}, which explores feature transformations between high and low resolutions, achieves sub-optimal results in both full-reference and no-reference metrics.
This outcome suggests that UHDDIP prioritizes restoring structure and detail, thereby improving the perceptual quality of images, whereas UHD focuses more on overall naturalness and aesthetics.
%%%%%%%%%%%%%%%%%%%%%%
\begin{table*}[t]
\setlength{\tabcolsep}{9.6pt}
\caption{Image deblurring results on the UHD-Blur~\cite{uhdformer} dataset. The \colorbox{pink!50}{\textbf{best}} and \colorbox{blue!10}{second-best} results for each metric are highlighted.}
\label{tab:image deblurring.} 
\centering
\renewcommand{\arraystretch}{1.3}
\begin{tabular}{l|l|c|ccc|ccc}
\shline
\textbf{Type}
&\textbf{Methods}
&\textbf{Venue}
&\textbf{~~PSNR~$\uparrow$~} 
&\textbf{SSIM~$\uparrow$~}
&\textbf{LPIPS~$\downarrow$~}
&\textbf{~~NIQE~$\downarrow$~} 
&\textbf{MUSIQ~$\uparrow$~}
&\textbf{PI~$\downarrow$~}
\\
\shline
\multirow{5}{*}{\textbf{Non-UHD}} 
&MIMO-Unet++~\cite{mimo-unet++}&ICCV’21&25.025  &0.7517  &0.3874 & 8.1240&	24.5888&	7.2059 \\
&Restormer~\cite{Zamir2021Restormer}&CVPR’22&25.210  &0.7522 &0.3695 & 7.9269&	 25.4992&	6.9449 \\
&Uformer~\cite{wang2021uformer}&CVPR’22&25.267 &0.7515 &0.3851 &7.9875&	23.8844&	7.0959  \\
&Stripformer~\cite{Stripformer}&ECCV’22&25.052 &0.7501 &0.3740 &8.2872&	25.1366&	7.3679 \\
&FFTformer~\cite{FFTformer}&CVPR’23&25.409 &0.7571 &0.3708 &\colorbox{pink!50}{\textbf{5.8325}}&	\colorbox{blue!10}{27.2380}&	 7.0416\\
\shline
\multirow{3}{*}{\textbf{UHD}} 
&UHDformer~\cite{wang2024uhdformer}&AAAI’24& 28.821 &0.8440 &\colorbox{blue!10}{0.2350}&6.4782&	 25.3655&\colorbox{pink!50}{\textbf{6.0820}} \\
&MixNet~\cite{MixNet}&ArXiv’24&\colorbox{blue!10}{29.43} &\colorbox{blue!10}{0.855} & -&-&-&-\\
&UHDDIP~\cite{uhddip}&ArXiv’24&\colorbox{pink!50}{\textbf{29.517}}&\colorbox{pink!50}{\textbf{0.8585}} &\colorbox{pink!50}{\textbf{0.2127}} &\colorbox{blue!10}{6.3416}&	\colorbox{pink!50}{\textbf{28.2130}}&	\colorbox{blue!10}{6.1956}  \\
%&TSFormer~\cite{tsformer}&ArXiv’24&\textbf{29.52} & \textbf{0.861} &\textbf{0.203} &-&-&- \\ 
%&UHDPromer&-&\textbf{29.527} &\textbf{0.858} &\underline{0.216} &6.252&27.577&	6.127 \\
\shline
\end{tabular}
\end{table*}
%%%%%%%%%%%%%%%%%%%%%%
\begin{figure*}[t]
\centering
\begin{center}
\begin{tabular}{c}
\hspace{-2mm}\includegraphics[width=1\linewidth]{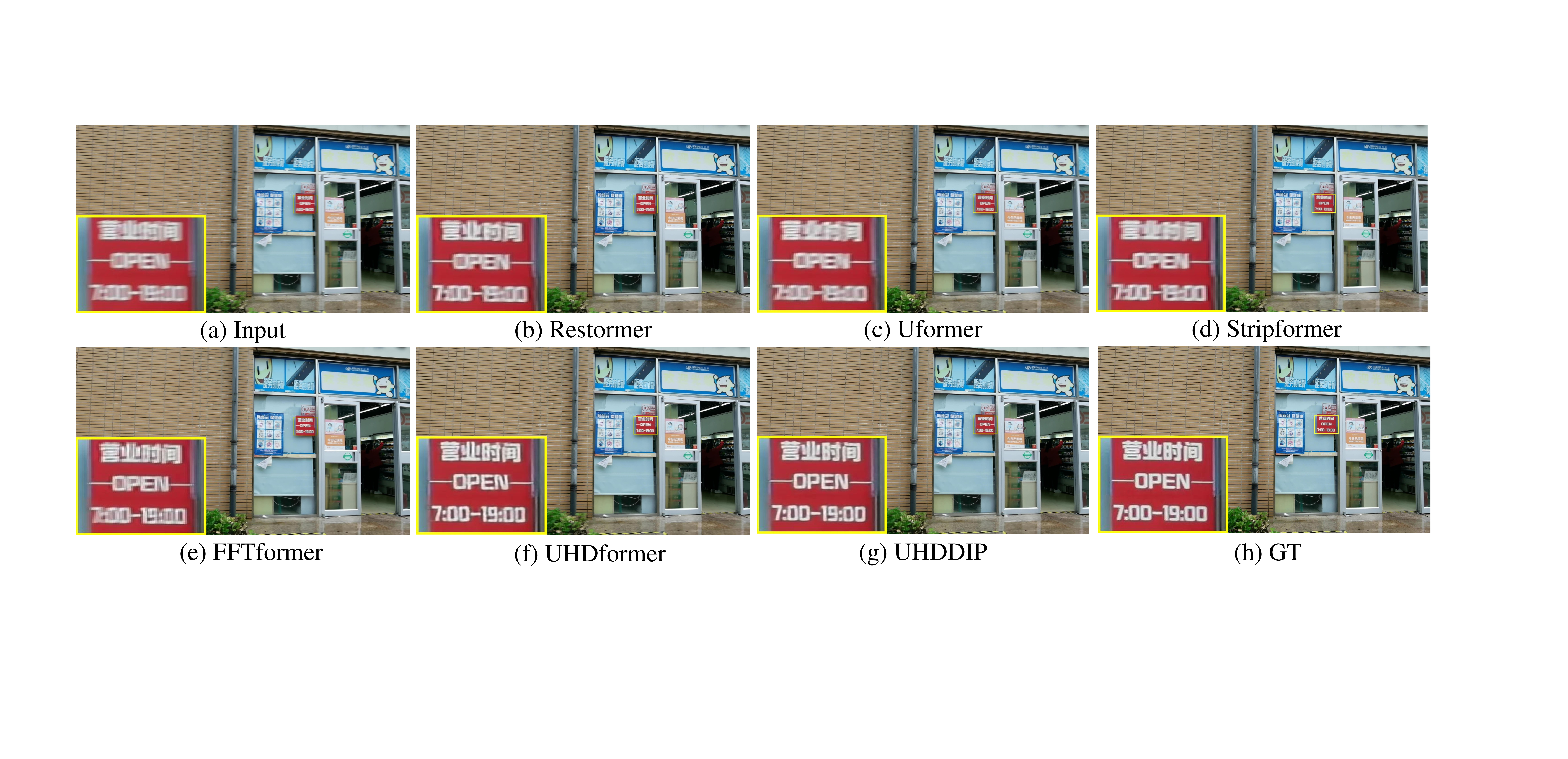}
\\
\end{tabular}
\caption{Visual results of different methods for UHD image deblurring on the UHD-Blur~\cite{uhdformer} dataset.}\label{fig:image deblurring on UHD-Blur}
\end{center}
\end{figure*}
\\
\\
\noindent\textbf{Qualitative Comparisons:}
We select a representative image from the UHD-Haze test results to evaluate the dehazed visualization, as shown in Fig.~\ref{fig:image dehazing on UHD-Haze}.
Restormer~\cite{Zamir2021Restormer} and Dehazeformer~\cite{DehazeFormer} fail to fully remove haze, leaving residual haze in the dehazed image. 
In addition, Uformer~\cite{wang2021uformer} and UHD~\cite{ZhengRCHWSJ21} introduce noticeable artifacts and blurring, detracting from the overall quality. In contrast, only UHDformer~\cite{uhdformer} and UHDDIP~\cite{uhddip} excel in effectively removing haze and delivering visually appealing results. 
However, when compared to the ground truth, UHDformer tends to over-enhance the image, introducing color bias. On the other hand, UHDDIP avoids over-enhancement, maintaining natural color fidelity, and resulting in outputs that closely align with the ground truth.

\subsubsection{UHD Image Deblurring}
For UHD image deblurring, we evaluate five non-UHD methods~\cite{mimo-unet++,Zamir2021Restormer,wang2021uformer,Stripformer,FFTformer} and three UHD methods, including UHDformer~\cite{uhdformer}, MixNet~\cite{MixNet}, and UHDDIP~\cite{uhddip} on the UHD-Blur dataset~\cite{uhdformer}.
%, which consists of 300 paired images.
%%%%%%%%%%%%%%%%%%%%
\\
\\
\noindent\textbf{Quantitative Comparisons:}
In addition to the MixNet results obtained from~\cite{MixNet}, we provide the PSNR, SSIM, and LPIPS metrics for the different methods in Table~\ref{tab:image deblurring.}, along with the calculated no-reference metrics for the recovered images.
Two key observations emerge. Firstly, UHDDIP~\cite{uhddip}, a prior interaction-based recovery method that incorporates gradient and normal priors of low-resolution features to guide high-resolution recovery, performs well on full-reference metrics and MUSIQ.
Secondly, methods like FFTformer~\cite{FFTformer} and UHDformer~\cite{uhdformer} show strong performance in terms of NIQE and PI by modeling global features. However, they are slightly inferior to MixNet in terms of PSNR and SSIM.
\\
\\
\noindent\textbf{Qualitative Comparisons:}
As shown in Fig.~\ref{fig:image deblurring on UHD-Blur}, the visualization results of different deblurring methods on the UHD-Blur dataset are displayed.
It is observed that all methods enhance the clarity of the input image to some extent. However, the deblurring performance of Restormer~\cite{Zamir2021Restormer}, Uformer~\cite{wang2021uformer}, and Stripformer~\cite{Stripformer} is relatively limited, while FFTformer~\cite{FFTformer} achieves better results but still falls short of producing fully clear images.
Conversely, UHDformer~\cite{uhdformer} and UHDDIP~\cite{uhddip} demonstrate the most significant advantages, ranking highest in performance. 
It is worth noting that UHDformer tends to introduce structural distortions and edge ghosting, whereas UHDDIP excels by preserving image texture and structure, achieving greater clarity without compromising the image's integrity.

\subsubsection{UHD Image Deraining $\&$ Desnowing}

\begin{figure*}[!th]
\centering
\begin{center}
\begin{tabular}{c}
\hspace{-2mm}\includegraphics[width=1\linewidth]{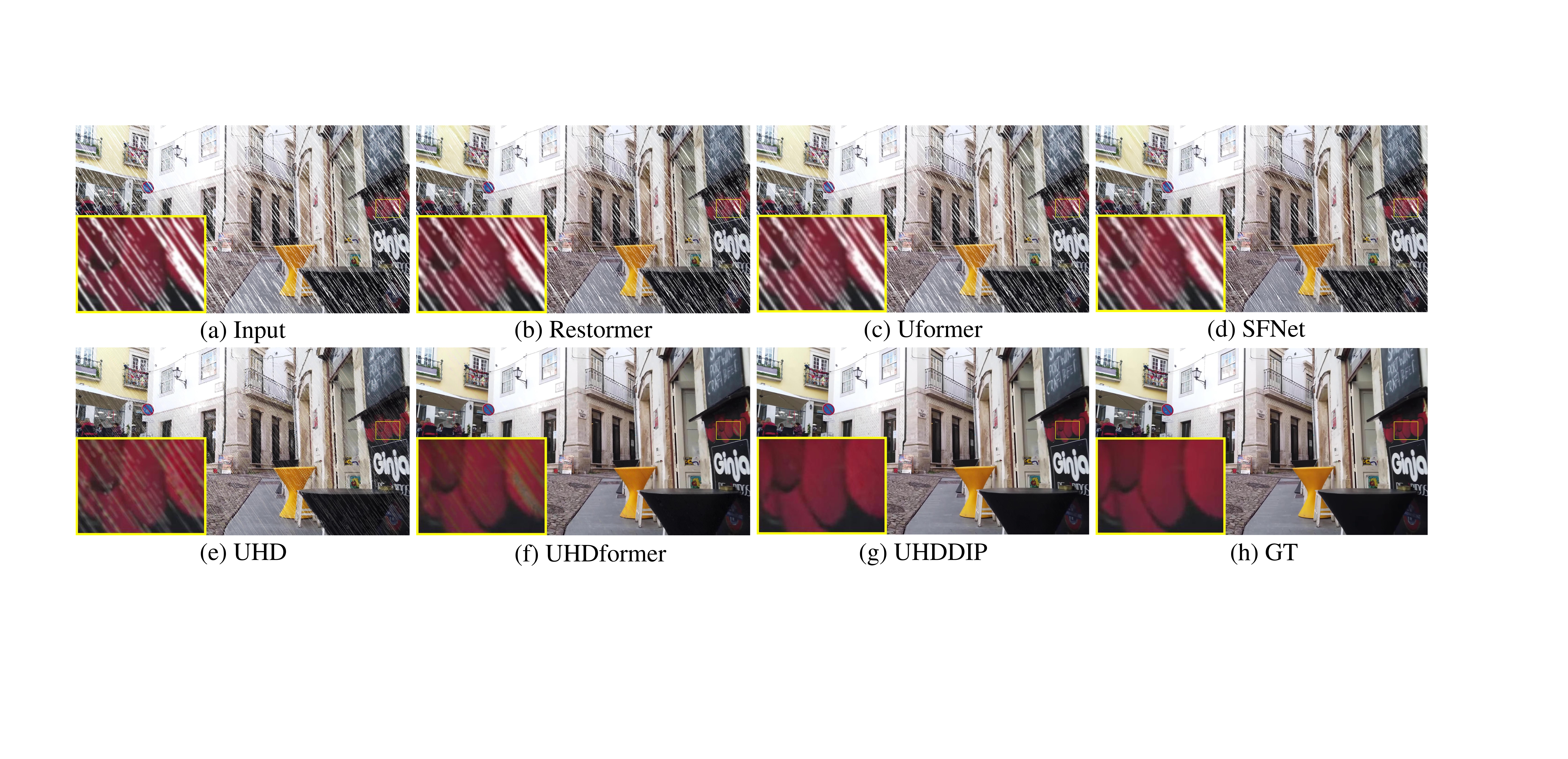}
\\
\end{tabular}
\caption{Visual results of different methods for UHD image deraining on the UHD-Rain~\cite{uhddip} dataset.}\label{fig:image deraining on UHD-Rain}
\end{center}
\end{figure*}

\begin{figure*}[!th]
\centering
\begin{center}
\begin{tabular}{c}
\hspace{-2mm}\includegraphics[width=1\linewidth]{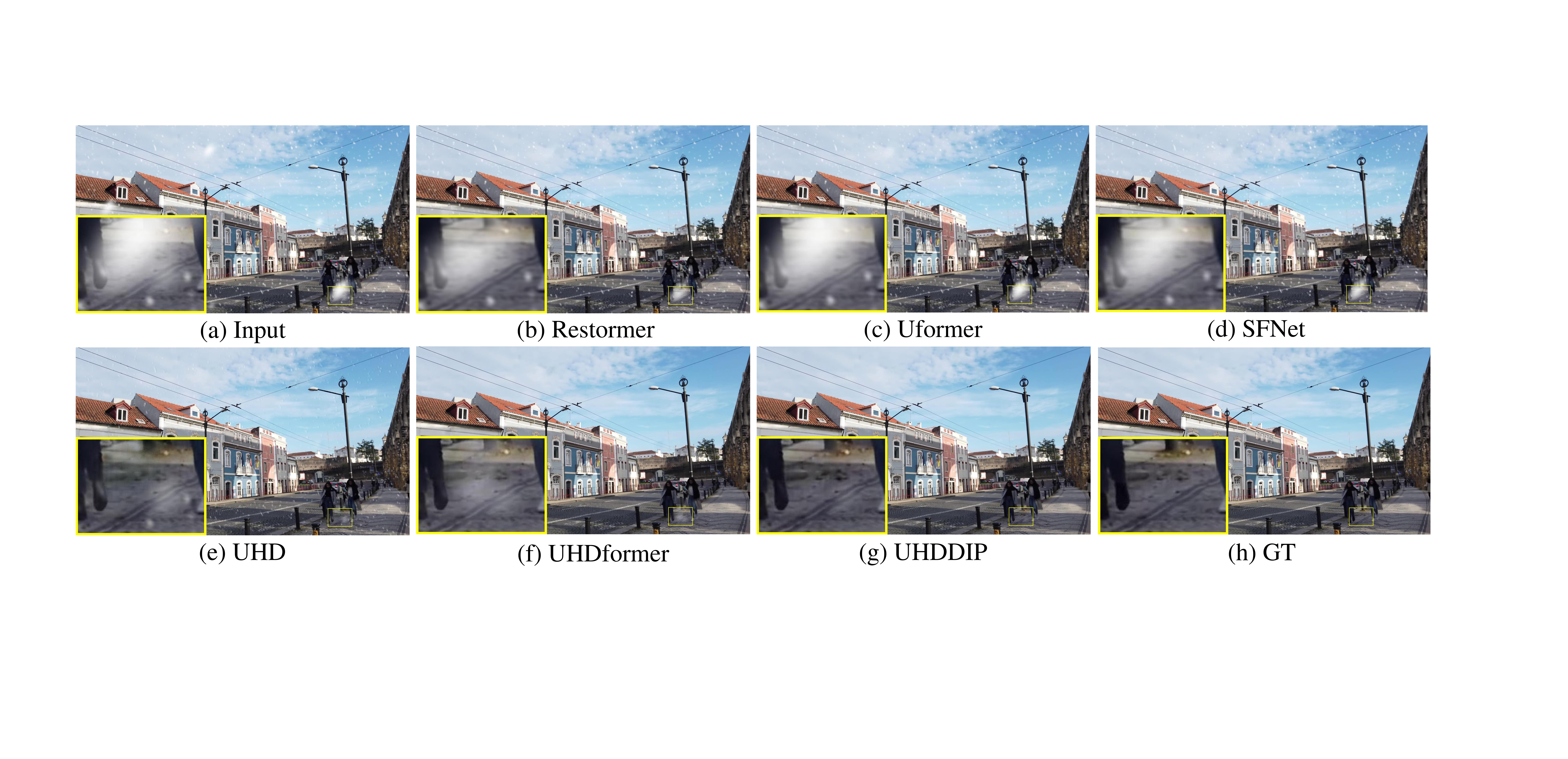}
\\
\end{tabular}
\caption{Visual results of different methods for UHD image desnowing on the UHD-Snow~\cite{uhddip} dataset.}\label{fig:image desnowing on UHD-Snow}
\end{center}
\end{figure*}

In this section, we compare several UHD image deraining and desnowing methods using the UHD-Rain and UHD-Snow datasets~\cite{uhddip}. The evaluation includes three non-UHD methods~\cite{Zamir2021Restormer,wang2021uformer,0001TBRGC0K23sfnet}, alongside three representative UHD image restoration methods~\cite{ZhengRCHWSJ21,uhdformer,uhddip}.
\\
\\
\noindent\textbf{Quantitative Comparisons:}
Table~\ref{tab:image deraining.} and Table~\ref{tab:image desnowing.} present the quantitative results of the UHD image deraining and desnowing methods, respectively. Six evaluation metrics—PSNR, SSIM, LPIPS, NIQE, MUSIQ, and PI—are used to quantitatively assess the performance of the different methods.
The results indicate that UHDDIP demonstrates competitive performance in terms of full-reference metrics, \ie, PSNR, SSIM, and LPIPS, for both deraining and desnowing tasks. This success is attributed to its effective utilization of prior feature interactions at low resolutions.
For no-reference metrics, UHDDIP achieves the best NIQE scores, highlighting superior perceptual quality. However, it lags slightly behind UHD and UHDFormer in terms of MUSIQ and PI scores for the desnowing task.
\\
\\
\noindent\textbf{Qualitative Comparisons:}
We provide qualitative comparisons of the various methods on the UHD-Rain and UHD-Snow datasets, as illustrated in Fig.~\ref{fig:image deraining on UHD-Rain} and Fig.~\ref{fig:image desnowing on UHD-Snow}. The non-UHD methods~\cite{Zamir2021Restormer,wang2021uformer,0001TBRGC0K23sfnet} demonstrate limited effectiveness, as they fail to adequately remove rain streaks and snowflakes.
Both UHD~\cite{ZhengRCHWSJ21} and UHDformer~\cite{uhdformer} exhibit the capability to reduce rain streaks and snowflakes but still leave behind some residual artifacts.
Specifically, UHD struggles with residual rain streaks and smaller snowflakes, while UHDformer is more effective at removing smaller snowflakes but remains less sensitive to rain streaks and larger snowflakes.
In contrast, UHDDIP~\cite{uhddip} excels by effectively eliminating both rain streaks and snowflakes while preserving the structure and finer details of the images, leading to superior overall performance.
%%%%%%%%%%%%%%%%%%%%%%%%%
\begin{table*}[t]
\setlength{\tabcolsep}{10.25pt}
\caption{Image deraining results on the UHD-Rain~\cite{uhddip} dataset. The \colorbox{pink!50}{\textbf{best}} and \colorbox{blue!10}{second-best} results for each metric are highlighted.}
\label{tab:image deraining.} 
\centering
\renewcommand{\arraystretch}{1.3}
\begin{tabular}{l|l|c|ccc|ccc}
\shline
\textbf{Type}
&\textbf{Methods}
&\textbf{Venue}
&\textbf{~~PSNR~$\uparrow$~} 
&\textbf{SSIM~$\uparrow$~}
&\textbf{LPIPS~$\downarrow$~}
&\textbf{~~NIQE~$\downarrow$~} 
&\textbf{MUSIQ~$\uparrow$~}
&\textbf{PI~$\downarrow$~}
\\
\shline
\multirow{3}{*}{\textbf{Non-UHD}} 
&Uformer~\cite{wang2021uformer}&CVPR’22&19.494 &0.7163 &0.4598& 9.9422 &	25.0728&	8.0579  \\
&Restormer~\cite{Zamir2021Restormer}&CVPR’22&19.408  &0.7105  &0.4775 &11.2731&	24.9277&	8.8409   \\
&SFNet~\cite{0001TBRGC0K23sfnet}&ICLR’23&20.091 &0.7092 &0.4768 &8.0167&	22.3553&	 7.1914 \\
\shline
\multirow{3}{*}{\textbf{UHD}} 
&UHD~\cite{ZhengRCHWSJ21}&ICCV’21&26.183 &0.8633 & 0.2885  &8.0167&	22.3553&	 7.1914\\
&UHDformer~\cite{wang2024uhdformer}&AAAI’24& \colorbox{blue!10}{37.348} &\colorbox{blue!10}{0.9748}&\colorbox{blue!10}{0.0554}  &\colorbox{blue!10}{4.9256}&	\colorbox{blue!10}{30.4390}&	\colorbox{blue!10}{5.0464}\\
&UHDDIP~\cite{uhddip}&ArXiv’24&\colorbox{pink!50}{\textbf{40.176}} &\colorbox{pink!50}{\textbf{0.9821}} &\colorbox{pink!50}{\textbf{0.0300}}  & \colorbox{pink!50}{\textbf{4.8796}}& \colorbox{pink!50}{\textbf{31.0163}}&	\colorbox{pink!50}{\textbf{5.0383}} \\
%&TSFormer~\cite{tsformer}&ArXiv’24&\textbf{40.40} & \textbf{0.983} &\textbf{0.028} &-&-&- \\
\shline
\end{tabular}
\end{table*}
%
%%%%%%%%%%%%%%%%%%%%%%
%%%%%%%%%%%%%%%%%%%%%%%%%

\subsection{Computational Complexity}\label{sec: Computational Complexity}
\noindent In Table~\ref{tab:Efficiency comparison}, we compare the computational complexity of non-UHD and UHD methods, including the number of network
model parameters, FLOPS (floating-point operations per second), and the run time for processing a single image. 
%
%These results are obtained on a single NVIDIA RTX2080Ti GPU, using a batch size of 1 at a resolution of $1024\times1024$.
For a fair comparison, we calculate the number of parameters and FLOPs with a batch size of 1 at a resolution of $1024 \times 1024$ pixels, using a single NVIDIA RTX2080Ti GPU. The official event function from CUDA is employed to measure the runtime.
Additionally, the results for TSFormer are obtained from~\cite{tsformer}, while the FLOPs and run time for NSEN~\cite{nsen} are not reported, as their codes are not publicly available.
The results clearly indicate that UHD methods outperform non-UHD methods in terms of the number of parameters, FLOPs, and average run time. Among the UHD methods, TSFormer~\cite{tsformer} achieves the lowest FLOPs and shortest run time. In contrast, UHDFormer~\cite{uhdformer} uses significantly fewer parameters compared to the other methods.
UHD~\cite{ZhengRCHWSJ21} ranks second in run time, but the use of three deep CNNs leads to a relatively higher number of parameters and FLOPs.
In addition, UHDDIP~\cite{uhddip} has slightly more parameters and FLOPs than the most efficient methods but maintains a moderate run time, suggesting its potential for practical applications.

\section{Future Research Directions}\label{Sec: Challenges and Opportunities}
Ultra-high-definition (UHD) image restoration represents a promising area of research, particularly due to the inherent challenges posed by the high computational complexity of UHD images. 
In this section, we explore potential future research directions for UHD image restoration, offering insights that may guide researchers in addressing existing challenges and uncovering novel solutions.

\subsection{Effective Processing Paradigm}
As previously discussed, current UHD image restoration methods predominantly follow two main processing paradigms: the downsampling-enhancement-upsampling and the encoder-decoder with stepwise up-downsampling. 
These paradigms aim to reduce computational costs by employing high magnification or stepwise downsampling; however, a significant drawback is the information loss which degrades overall restoration quality.
Although some researchers have attempted to avoid information loss through wavelet transforms~\cite{Wave-Mamba}, the connection with the augmentation network has not been considered, and its effectiveness requires further improvement. 
Additionally, advancements in learning model-aware resampling methods~\cite{LMAR} have shown potential for preserving feature consistency between UHD image inputs and their corresponding low-resolution inputs without requiring retraining of the enhancement network.
Thus, exploring the intrinsic connection between the resampling operation and the enhancement network is a potential research direction for developing more effective processing paradigms.
%%%%%%%%%%%%%%%%%%%
\begin{table*}[t]
\setlength{\tabcolsep}{10.25pt}
\caption{Image desnowing results on the UHD-Snow~\cite{uhddip} dataset. The \colorbox{pink!50}{\textbf{best}} and \colorbox{blue!10}{second-best} results for each metric are highlighted.}
\label{tab:image desnowing.} 
\centering
\renewcommand{\arraystretch}{1.3}
\begin{tabular}{l|l|c|ccc|ccc}
\shline
\textbf{Type}
&\textbf{Methods}
&\textbf{Venue}
&\textbf{~~PSNR~$\uparrow$~} 
&\textbf{SSIM~$\uparrow$~}
&\textbf{LPIPS~$\downarrow$~}
&\textbf{~~NIQE~$\downarrow$~} 
&\textbf{MUSIQ~$\uparrow$~}
&\textbf{PI~$\downarrow$~}
\\
\shline
\multirow{3}{*}{\textbf{Non-UHD}} 
&Uformer~\cite{wang2021uformer}&CVPR’22&23.717 & 0.8711 &0.3095 &8.5964&26.1323&7.5036  \\
&Restormer~\cite{Zamir2021Restormer}&CVPR’22&24.142  & 0.8691  &0.3190& 9.4711&26.0491&	8.0490      \\
&SFNet~\cite{0001TBRGC0K23sfnet}&ICLR’23&23.638 &0.8456 &0.3528 &  7.3665&	22.5295&	 7.1527\\
\shline
\multirow{3}{*}{\textbf{UHD}} 
&UHD~\cite{ZhengRCHWSJ21}&ICCV’21&29.294 &0.9497 &0.1416 &4.8127&	 \colorbox{pink!50}{\textbf{33.2400}}&	5.0815 \\
&UHDformer~\cite{wang2024uhdformer}&AAAI’24&\colorbox{blue!10}{36.614} &\colorbox{blue!10}{0.9881} &\colorbox{blue!10}{0.0245}   &\colorbox{blue!10}{4.7680}&	31.1335&	\colorbox{pink!50}{\textbf{5.0046}}\\
&UHDDIP~\cite{uhddip}&ArXiv’24&\colorbox{pink!50}{\textbf{41.563}}&\colorbox{pink!50}{\textbf{0.9909}} &\colorbox{pink!50}{\textbf{0.0179}}  & \colorbox{pink!50}{\textbf{4.7659}}&	\colorbox{blue!10}{31.2507}&\colorbox{blue!10}{5.0048}  \\
%&TSFormer~\cite{tsformer}&ArXiv’24&\textbf{41.82} & \textbf{0.992} &\textbf{0.016} &-&-&- \\
\shline
\end{tabular}
\end{table*}
%
%%%%%%%%%%%%%%%%%%%%%%
%%%%%%%%%%%%%%%%%%%%%%%%%
%%%%%%%%%%%%%%%%%%%%%%%%%%
\begin{table*}[t]
\centering
\renewcommand{\arraystretch}{1.3}
\setlength{\tabcolsep}{24.5pt}
\caption{Efficiency comparison, 
reporting the number of parameters, FLOPs, and run time. 
The testing is conducted on a single RTX2080Ti GPU with a batch size of 1 at a resolution of $1024 \times 1024$ pixels.
The \colorbox{pink!50}{\textbf{best}} and \colorbox{blue!10}{second-best} results for each metric are highlighted.
}
\label{tab:Efficiency comparison} 
\renewcommand{\arraystretch}{1.3}
\begin{tabular}{l|c|ccc}
\shline
 \textbf{Methods} & \textbf{Type}  &\textbf{Parameters (M)~$\downarrow$} &\textbf{FLOPs (G)~$\downarrow$} 
 & \textbf{Run Time (s)~$\downarrow$}
\\
\shline
Restormer~\cite{Zamir2021Restormer}& \multirow{6}{*}{\textbf{Non-UHD}}&26.10 &2255.85 & 1.86 \\
Uformer~\cite{wang2021uformer}&& 20.60 &657.45 &0.60\\
SFNet~\cite{0001TBRGC0K23sfnet}&&13.23 &1991.03 &0.61\\
DehazeFormer~\cite{DehazeFormer}&&2.51 &375.40 &0.45\\
Stripformer~\cite{Stripformer}&&19.71 &2728.08 &0.15\\
FFTformer~\cite{FFTformer}&&16.56 &2104.60 &1.27\\
\shline
LLFormer~\cite{LLformer}& \multirow{10}{*}{\textbf{UHD}}&13.13 &221.64 & 1.69\\
UHD~\cite{ZhengRCHWSJ21}&&34.55  &113.45 &0.04 \\
UHDFour~\cite{LiGZLZFL23}&& 17.54  &75.63 &\colorbox{blue!10}{0.02}  \\
NSEN~\cite{nsen}&&2.67 &- &-  \\
UHDformer~\cite{wang2024uhdformer}&& \colorbox{pink!50}{\textbf{0.34}}&48.37 &0.16\\
Wave-Mamba~\cite{Wave-Mamba}&&1.26 &119.06 &0.66   \\
%DMixer~\cite{dmixer}&&16.32 &0.651 &   \\
MixNet~\cite{MixNet}&&7.77 &463.69 &0.14  \\
UHDDIP~\cite{uhddip}& &\colorbox{blue!10}{0.81} &\colorbox{blue!10}{34.73}  &0.13 \\
UDR-Mixnet~\cite{UDRMixer}& &4.87 &199.96  &0.12 \\
TSFormer~\cite{tsformer}& &3.38 &\colorbox{pink!50}{\textbf{24.73}}  &\colorbox{pink!50}{\textbf{0.01}} \\
\shline
\end{tabular}
\end{table*}

\subsection{Lightweight Network Design}
Existing deep learning models face significant computational challenges, which limit their application in UHD image restoration tasks. 
Current restoration methods, ranging from traditional CNNs to advanced Transformer architectures, demand substantial computational power, making them heavily reliant on high-end GPUs. 
Although some networks have been proposed to reduce these complexities through MLPs and novel Mamba frameworks, they fall short of enabling direct implementation on mobile devices. Consequently, future research should prioritize the development of lightweight network architectures that balance performance and efficiency, enabling broader applicability across diverse devices.

\subsection{Developing Real-World Benchmark Datasets}
As shown in Table~\ref{tab:Benchmark Datasets}, except for the UHD-LL dataset~\cite{LiGZLZFL23}, most existing benchmark datasets are artificially synthesized. 
Although the data shortage in the field of UHD image restoration has been temporarily alleviated, a significant gap remains between synthetic images and real-world degraded images. 
Models trained on synthetic images often perform well on synthetic test samples but exhibit poor performance on real-world images. For this reason, it is imperative to develop large-scale paired training datasets comprising real-world UHD images.

\subsection{Using Image Priors}
Current UHD image restoration networks predominantly focus on learning intrinsic features; however, the incorporation of image priors plays a pivotal role in enhancing image quality.
For instance, UHDDIP~\cite{uhddip} incorporates gradient and normal priors into its model design, significantly improving the structural integrity and detail preservation of restored images. This demonstrates the potential of leveraging image priors to guide the restoration process.
Further exploration of alternative image priors opens promising avenues for advancing UHD image restoration methodologies.

\subsection{Specialized Evaluation Metrics}
Most image quality evaluation metrics, such as PSNR, SSIM, and LPIPS, are designed for standard-resolution images and reflect image quality to a certain extent.
However, the high-resolution characteristics of UHD images place greater demands on detail restoration and subjective visual perception. 
For example, the perceived quality of an image may be affected by local detail in different areas, which may not be evident in low-resolution images. 
Therefore, it is crucial to develop evaluation indicators specifically tailored for UHD images. These metrics should accurately reflect the unique requirements of UHD images in terms of detail representation and overall sensory quality.

\subsection{UHD Images with Multiple Degradations}
Current UHD image restoration algorithms are usually designed to address a single type of degradation. However, in practical applications, UHD images are often affected by a combination of degradation factors, leaving the need for mixed degradation processing largely unresolved. It is worth noting that SimpleIR~\cite{SimpleIR} is the first to propose an all-in-one restoration method for UHD images.
While all-in-one image restoration methods have shown significant progress on low-resolution images using prompt learning and dynamic network technologies, these techniques have not yet been applied to UHD images.

\section{Conclusion}\label{Sec: Conclusion}
In this paper, we present a comprehensive survey of recent advancements in deep learning-based UHD image restoration methods, making the first systematic survey in this domain. We have explored key components, including degradation models, benchmark datasets, and the primary strategies employed for various UHD restoration tasks.
To effectively organize the existing approaches, we classify them into three primary categories based on their processing strategies: (1) downsampling-enhancement-upsampling, (2) encoder-decoder with stepwise up-downsampling, and (3) resampling-enhancement.
Furthermore, we analyze these methods from a practical perspective, providing both qualitative and quantitative comparisons across evaluation metrics and related tasks. 
While significant progress has been made in this field, several promising directions \ie, efficient processing paradigms, lightweight network architectures, real-world benchmark datasets, advanced image priors, specialized evaluation metrics, and UHD image restoration methods capable of addressing multiple degradations. 
We hope this paper serves as a valuable resource for both beginners and professionals in the field of UHD image restoration and inspires continued innovation in this field.

% \clearpage
% ---- Bibliography ----
%
% BibTeX users should specify bibliography style 'splncs04'.
% References will then be sorted and formatted in the correct style.
%
% \bibliographystyle{splncs04}
% \bibliography{egbib}
\bibliographystyle{IEEEtran}
% \bibliography{aaai23}
\bibliography{egbib}
\end{document}